\documentclass{article}

\usepackage[preprint]{neurips_2026}

\usepackage[utf8]{inputenc} 
\usepackage[T1]{fontenc}    
\usepackage{hyperref}       
\usepackage{url}            
\usepackage{booktabs}       
\usepackage{amsfonts}       
\usepackage{nicefrac}       
\usepackage{microtype}      
\usepackage[table]{xcolor}         

\usepackage{amsfonts}
\usepackage{amsmath}
\usepackage{amssymb}
\usepackage{amsthm}
\usepackage{mathtools}
\usepackage{bm}
\usepackage{booktabs}
\usepackage{makecell}
\usepackage{subcaption}
\usepackage{tikz}
\usepackage{multirow}
\usepackage{colortbl}
\usepackage{longtable}
\usepackage{xspace}
\usepackage{pgf}
\usepackage{enumerate}   
\usepackage{bookmark}

\newcommand{\simon}[1]{}
\newcommand{\michael}[1]{}
\newcommand{\martin}[1]{}
\newcommand{\todo}[1]{}
\newcommand*{\eg}{e.g.\@\xspace}
\newcommand*{\ie}{i.e.\@\xspace}

\newlength{\firstword}

\newlength{\firstwordagain}

\definecolor{covred}{rgb}{0.81, 0.09, 0.13}
\definecolor{covgreen}{rgb}{0.13, 0.55, 0.13}
\definecolor{covbg}{RGB}{250,250,250}

\makeatletter
\newcommand{\btcoloredrowstrut}{%
  \rule[-\dimexpr\dp\@arstrutbox+\aboverulesep\relax]{0pt}%
        {\dimexpr\ht\@arstrutbox+\dp\@arstrutbox+\aboverulesep+\belowrulesep\relax}%
}
\makeatother

\definecolor{covred}{HTML}{D62728}
\definecolor{covgreen}{HTML}{1B8E2B}
\newcommand{\coverage}[2]{%
  \begingroup
  \pgfmathtruncatemacro{\isfull}{ifthenelse(#2>=99.5,1,0)}%
  \ifnum\isfull=1
    \textcolor{covgreen!#2!covred}{\textbf{#1 (#2)}}%
  \else
    \textcolor{covgreen!#2!covred}{#1 (#2)}%
  \fi
  \endgroup
}

\title{Efficient Lookahead Encoding and Abstracted Width for Learning General Policies in Classical Planning}
\author{
    Michael Aichm\"uller\\
    RWTH Aachen University\\
    Aachen, Germany \\
    \texttt{michael.aichmueller@ml.rwth-aachen.de}\\
    \And
    Simon St\r{a}hlberg \\
    RWTH Aachen University\\
    Aachen, Germany \\
    \texttt{simon.stahlberg@gmail.com}\\
    \And
    Martin Funkquist\\
    Link\"{o}ping University\\
    Link\"{o}ping, Sweden \\
    \texttt{martin.funkquist@liu.se}\\
    \And
    Hector Geffner\\
    RWTH Aachen University\\
    Aachen, Germany \\
    \texttt{hector.geffner@ml.rwth-aachen.de}
}

\begin{document}

\maketitle

\begin{abstract}
    Generalized planning aims to learn policies that generalize across large collections of instances within a classical planning domain. Recent approaches using Graph Neural Networks (GNNs) have shown the ability to learn nearly perfect policies for several domains. This work improves on the recently published idea of Iterated Width (IW) policies. Therein, the policy broadens its successor scope through an IW-lookahead search that offers to ``jump'' over multiple transitions, simplifying the problem structure. Yet, each transition is evaluated individually leading to unscalable compute and expressivity limitations. Furthermore, while an IW(1) search is attractive due to scaling linearly with the number of atoms in a problem, it still becomes inefficient once thousands of objects are considered like in the International Planning Competition (IPC) 2023 benchmark. In this work, we address both limitations. Firstly, we introduce a vastly more efficient holistic encoding of the entire search tree. It jointly represents IW(1)-reachable states only by their relational differences to the current state, which enables Relational GNNs (R-GNNs) to score all transitions in a single forward pass.  
Secondly, we define Abstracted IW(1) to improve scaling through relational abstraction during novelty checks. Rather than testing fully instantiated atoms, it abstracts each atom by replacing all but one of its arguments with their types. The original atom is then novel if any of its abstracted forms is. This structural compression shifts the scaling of the novelty search to be linear in the number of objects, rather than atoms, still capturing meaningful subgoal structure. 
Our contributions are evaluated on the hyperscaling IPC 2023 benchmark and across a broad range of domains, including those that require features beyond the $C_2$ logic fragment. The results show that our policies achieve a new state-of-the-art performance, significantly surpassing prior work, including classical planners like LAMA.

\end{abstract}

\section{Introduction}
\label{sec:introduction}

Generalized planning aims to learn policies that generalize across large
collections of planning problems within the same domain, differing only in
initial states, goals, and object counts
\citep{srivastava-et-al-aij2011,hu-degiacomo-ijcai2011,jimenez-et-al-ker2019}. A
general policy allows for solving diverse planning problems greedily,
eliminating the need for search. Various approaches have been pursued to learn
such policies from domain descriptions and training instances, ranging from
symbolic to deep learning methods.
While symbolic approaches offer transparency and potential correctness
guarantees, they often struggle with scalability
\citep{khardon-aij1999,martin-geffner-ai2004,fern-et-al-jair2006,srivastava-et-al-aij2011,boutilier-et-al-ijcai2001,sanner-boutelier-aij2009,frances-et-al-aaai2021}.
Conversely, deep learning approaches, though opaque, tend to scale more
effectively
\citep{toyer-et-al-jair2020,bajpai-et-al-neurips2018,rivlin-et-al-icaps2020wsprl,stahlberg-et-al-kr2023}.

State-of-the-art generalized planning methods include supervised
transformer-based approaches \citep{rossetti-et-al-icaps2024} and unsupervised
reinforcement learning (RL) approaches using graph neural networks (GNNs),
specifically relational GNNs.
Both paradigms achieve impressive results on benchmarks, despite the known
expressivity limitations of GNNs and transformers
\citep{xu-et-al-iclr2019,barcelo-et-al-iclr2020,grohe-lics2021}.

Recently, \citet{aichmueller-geffner-ijcai2025} introduced IW-policies, which use the Iterated-Width (IW) algorithm \citep{lipovetzky-geffner-ecai2012} to perform lookahead searches at each state to augment successor choices. 
This enables ``jumps'' over multiple transitions, often simplifying the problem structure and reducing expressivity requirements.
These are integrated by replacing 1-step transitions $(s,s')$ with multi-step transitions $[s,s']$ identified by IW(1) from $s$ \citep{aichmueller-geffner-ijcai2025}. 
However, the embeddings for each object $o$ were inferred independently for states $s$ and $s'$, interacting only within the value-function readout. 
While feasible for small instances, this approach faces key \emph{computational and expressivity limitations} as size increases, because a given state might have many successors, of which each receives a separate and complete encoding of $s'$.

Our primary contributions address scaling and expressivity through two key innovations.
    First, we introduce a novel joint encoding approach for an entire tree of state transitions $(s,s')$, which avoids wastefully re-encoding shared state descriptions. We propose processing atoms from both $s$ and each $s'$ in an \emph{aggregated} encoding within the R-GNN. By recreating the lookahead tree structure within the graph-encoding and representing successors merely by their atom differences to the current state, we compute all values in a single pass, reducing memory usage by over a magnitude. 

    Second, we propose Abstracted IW($1$), a significantly relaxed variant of IW($1$). For novelty testing, this variant abstracts state atoms by iteratively replacing all but one argument with their most specific type and tests the abstracted atoms instead. This shifts the worst-case complexity of the search from being linear in the number of atoms to the number of objects. 
          Consequently, Abstracted IW($1$) evaluates significantly faster while still capturing meaningful subgoals for task decomposition. 
          Furthermore, this lookahead mitigates the expressivity limitations of models restricted to immediate successors; it simplifies the problem's cost structure and removes the need for features the R-GNN inherently lacks the expressivity to derive.

These improvements are essential for scaling to large state spaces with numerous successors, a necessity highlighted by the massive instances in the IPC 2023 Learning Track (e.g., a 488-block Blocksworld). 
Scaling to this level requires both highly efficient IW lookahead and excellent model generalization. 
By achieving this, our approach outperforms prior work on the benchmark, including the strong LAMA baseline. 
Finally, the proposed lookahead drives substantial improvements in domains requiring expressivity beyond the standard $C_2$ logic fragment \cite{drexler-et-al-kr2024}.

The paper is organized as follows. We begin with background on classical and generalized planning, and IW, including its use for state-space jumps. We then introduce R-GNNs, followed by the input encodings and the reinforcement learning training procedure. Finally, we present the experimental evaluation, discuss the results, and conclude. Related work is discussed throughout the paper.

\section{Planning}
\label{sec:planning}

\paragraph{Classical Planning.}
A classical planning instance is a pair $P=\langle D,I \rangle$, where
$D=\langle \mathcal{T}, \mathcal{P}, \mathcal{A} \rangle$ is the domain and $I=\langle
\mathcal{O}, s_0, g \rangle$ specifies the instance. The domain $D$ consists of
a set of types $\mathcal{T}$ organized in a hierarchy, along with a set of predicates $\mathcal{P}$ and actions $\mathcal{A}$. A predicate $P \in
\mathcal{P}$ has an arity $n$ and an associated tuple of types $(\tau_1, \dots, \tau_n) \in \mathcal{T}^n$, and an atom $P(x_1, \dots, x_n)$ is formed by
applying $P$ to a tuple of terms $(x_1, \dots, x_n)$ of matching types. If all terms are objects
from $\mathcal{O}$, the atom is said to be ground. Actions in $\mathcal{A}$ are
defined by a name and a typed list of arguments. Each action is associated with a
precondition (a set of atoms required to be true) and an effect (sets of atoms
to be added or deleted). We denote the set of actions applicable in a state $s$
as $\mathcal{A}[s]$. The instance $I$ defines the set of objects $\mathcal{O}$,
where each object $o \in \mathcal{O}$ is assigned a type $\tau \in \mathcal{T}$,
the initial state $s_0$, and the goal $g$, where states are sets of ground
atoms. A state $s$ is a goal state if $g \subseteq s$. A plan is a sequence of
applicable actions that transforms $s_0$ into a goal state.

\paragraph{Generalized Planning.}

A generalized planning problem is a set $\mathcal{Q}$ of planning instances $P$
drawn from the same domain, possibly with different object sets, initial states,
and goals. Unlike classical planning, where a solution is typically an open-loop
sequence of ground actions, a solution to a generalized problem is a closed-loop
policy $\pi$ that selects actions based on the current state and goals. This
policy must work for every instance in $\mathcal{Q}$. Formally, $\pi$
is correct for $\mathcal{Q}$ if for every $P \in \mathcal{Q}$ the trajectory
obtained by repeatedly applying $\pi$ from the initial state reaches a goal
state. Generalized planning thus focuses on compact, reusable control policies
that generalize across instances.

\section{Iterated Width}
\label{sec:iw}

Iterated Width (IW) is a search algorithm that prunes states based
on their \emph{novelty} \citep{lipovetzky-geffner-ecai2012}. For a state $s$
generated in the search, the novelty $w(s)$ is defined as the size of the
smallest tuple of atoms true in $s$ that is false in all previously generated
states. If no such tuple exists, the novelty is infinite. IW($k$) is a standard
Breadth-First Search (BFS) that prunes any state $s$ with novelty $w(s) > k$.
Since the number of atom tuples of size $k$ is polynomial in the number of
ground atoms, IW($k$) explores a polynomial number of states. Specifically,
IW(1) runs in time linear in the number of reachable atoms and preserves
solvability for problems with width 1.

IW is particularly effective because atomic goals in many classical planning domains exhibit a small width, meaning goals can be reached via sequences of states with persistently low novelty, regardless of problem size.
For example, in the Delivery domain, picking up $n$ packages and delivering them requires exactly $2n$ IW(1) jumps, regardless of grid size or package locations, as each subgoal is achievable within two jumps\footnote{Width-2 goals cannot always be decomposed into width-1 subgoals solvable by IW(1) \citep{bonet-geffner-jair2024}.}.
Consequently, IW provides a guarantee that blind search methods, such as bounded-depth BFS, fundamentally lack. 

In this work, we use IW(1) as a local lookahead mechanism to define
``jumps'' in the state space. An IW(1)-jump from $s$ leads to a state $s'$ that
is reachable from $s$ using IW(1). This allows the learned policy to reason only over the endpoints of a subproblem that can be solved efficiently by a low-complexity search,
focusing learning on high-level decisions. This approach has been explored
previously by
\cite{aichmueller-geffner-ijcai2025}. An important benefit of IW(1)-jumps
is that they often reduce the expressive requirements for learning the Q-value
function. This is because IW(1) can solve certain subproblems that would
otherwise require more complex features to represent, thus facilitating the
learning of a generalized policy.

\paragraph{Abstracted IW($k$)}
Each $n$-ary predicate $P(x_1, \dots, x_n)$ grounds
$O(|\mathcal{O}|^n)$ many atoms. Thus, although IW($1$) is linear in
the number of atoms, the number of atoms itself scales polynomially with
the object count. This bottleneck materializes in very large instances
and cannot be resolved through straightforward parallelization alone.
To avoid this polynomial scaling, we relax width-$1$ novelty by modifying
the novelty calculation itself.

For an $n$-ary ground atom $P(o_1, \dots, o_n)$ \emph{Abstracted IW($1$)} (AIW($1$)) iterates over each position $i$ to compress the information content of all but position $i$'s object slots via a reduction function $r(o)$.
The original atom is considered novel if any of its abstracted
atoms $P(r(o_1), \dots, r(o_{i-1}), o_i, r(o_{i+1}), \dots, r(o_n))$ is novel. In this framework, regular IW(1) corresponds to object identity
$r(o) = o$, whereas AIW(1) replaces
objects by their type in the planning domain,
\ie, $r(o) = \texttt{type}(o)$. In domains without a typing hierarchy,
AIW treats every object as belonging to the same universal type.

With this modification, the novel atom count is bounded by $O(t^{m-1} |\mathcal{O}|)$, where $t$ is the number of object types and $m$ is the maximum predicate arity. Since both $t$ and $m$ depend only on the domain, the number of novel atoms scales linearly in the problem's object count.
However, goal atoms may require full object identities to be
preserved to be safely detected in the lookahead. AIW therefore does not
abstract an atom $P(o_1, \dots, o_n)$ if it is mentioned in the goal.

The idea of AIW($1$) extends naturally to a general width-$k$ formulation.
Let $\ell = (q_i)_{i \in I}$ be a canonically ordered atom-tuple of
size $|I| \leq k$, and let $\mathcal{F}(q_i)$ denote the complete set of
abstracted forms of atom $q_i$ as described above. Then $\ell$ has
\emph{abstracted novelty at most $k$} if there exists a non-empty index set
$J \subseteq I$ such that the abstracted tuple $(a_j)_{j \in J}$
with  $a_j \in \mathcal{F}(q_j)$ has not been seen before in the search. 
Equivalently, AIW($k$) replaces the original subtuples of $\ell$ considered by
standard IW($k$) novelty with all abstracted tuples induced by $\ell$,
and marks $\ell$ novel if any abstracted tuple is.

Although AIW($k$) loses the completeness guarantees of IW($k$)
for problems of width $k$, it yields an efficiently scaling search that
still induces meaningful subproblem structure. In this paper, we focus on
AIW($1$), which allows us to overcome the branching hurdles of even
the largest test instances in several IPC domains. Unless stated otherwise
in the remainder, IW and AIW refer to IW($1$) and AIW($1$), respectively\footnote{We also considered a capacity-limited variant (C-AIW) that restricts the retained states per depth, and a version that forces object abstraction to a universal type in Base-Abstracted IW (BAIW). 
Details and results on these variants are in the appendix.}.

\section{Relational Graph Neural Networks}
\label{sec:rgnn}

While standard Graph Neural Networks (GNNs) operate on graphs, planning states are relational structures involving predicates of varying arities. Relational GNNs (R-GNNs) as used in Planning \citep{stahlberg-et-al-icaps2022} adapt standard message-passing GNNS to such structures by treating ground atoms as message sources that communicate information to the objects they involve. 
Indeed, by accommodating non-symmetric relations of arbitrary arity, R-GNNs generalize the standard homogeneous graph setting, where relations are typically represented as symmetric binary edges of a single type. Although such relations can be compiled away to obtain an equivalent GNN representation \citep{drexler-et-al-kr2024}, R-GNNs provide a higher-level abstraction that directly matches the relational state representations used in classical planning.

Concretely, each object $o\in\mathcal{O}$ has an embedding
$\textrm{f}_i(o)\in\mathbb{R}^k$ at iteration $i$, initialized to
$\textrm{f}_0(o)=\mathbf{0}$. An R-GNN takes as input a set of atoms, commonly a
state $s$ extended with the goal atoms, and produces final embeddings
$\textrm{f}_L(o)$, where $L$ is the number of R-GNN layers. These final
embeddings are then used to define the value or Q-functions that implicitly
encode the policies.

Let $\mathcal{R}$ be a set of ground atoms. Then the embeddings are updated by
message aggregation and updates for a fixed number of iterations as:
\begin{equation*}
\textrm{f}_{i+1}(o) \;=\; \textrm{f}_{i}(o) \;+\; \textrm{comb}_{U}\bigl(\textrm{f}_i(o),\,\textrm{m}_{o}\bigr),
\label{eq:rgnn:update:new}
\end{equation*}
where $\textrm{m}_o$ is the aggregated message received by $o$ from all atoms
that mention it. Each ground atom $q=P(o_1,\dots,o_n)\in\mathcal{R}$ computes a
set of position-wise messages $\{\textrm{m}_{q,o_j}\}_{j=1}^n$ by applying a
predicate-specific function $\textrm{comb}_P$ to the embeddings of the
participating objects:
\begin{equation*}
\big(\textrm{m}_{q,o_1},\dots,\textrm{m}_{q,o_n}\big)
\;=\;
\textrm{comb}_P\!\big(\textrm{f}_i(o_1),\dots,\textrm{f}_i(o_n)\big).
\label{eq:rgnn:msg:new}
\end{equation*}
In other words, $\textrm{comb}_P$ maps the tuple of object embeddings for atom
$q$ to a tuple of message vectors, one per argument position. The aggregated
message for object $o$ collects all such position-wise messages over atoms that
contain $o$:
\begin{equation*}
\textrm{m}_{o} \;=\; \mathrm{agg}\bigl(\{\,\textrm{m}_{q,o} \;:\; q\in\mathcal{R},\; o\in q\,\}\bigr).
\label{eq:rgnn:agg:new}
\end{equation*}
Here, it is important that $\mathrm{agg}(\cdot)$ is a permutation-invariant
function (\eg, maximization or summation).

We use the implementation provided by \citeauthor{stahlberg-geffner-aaai2026}
(\citeyear{stahlberg-geffner-aaai2026}), which uses maximization as the
aggregation function and multi-layer perceptrons (MLPs) as the combination
functions. Each MLP consists of two hidden layers with Mish activations
\citep{misra-bmva2020}.

\paragraph{Expressivity.}
The expressive power of GNNs is tightly linked to the Weisfeiler-Lehman (WL)
test \citep{morris-et-al-aaai2019,xu-et-al-iclr2019} and $C_2$ logic
\citep{cai-furer-immerman-combinatorica1992,barcelo-et-al-iclr2020,grohe-lics2021}.
This expressivity can be increased by defining vertices as $k$-tuples, yielding
$k$-GNNs that match the power of $k$-WL and $C_k$ logic for $k > 2$. However,
these more expressive architectures are data-hungry and difficult to train.
While architectures with $C_3$ power appear sufficient for most planning
benchmarks \citep{drexler-et-al-kr2024}, recent work suggests that theoretical
expressiveness does not always translate to performance. For instance, Edge
Transformers, which possess $C_3$ power
\citep{bergen-et-al-neurips2021,muller-et-al-neurips2024-transformers}, often
underperform in non-$C_2$ domains compared to more efficient
architectures that do not aim to capture the full expressivity of $C_3$
\citep{stahlberg-et-al-aaai2025}. 

\section{Transition Encodings in R-GNNs}
\label{sec:encodings}

Precisely how states and transitions are encoded as inputs to R-GNNs affects the
expressiveness of the resulting architecture.
\citeauthor{stahlberg-et-al-aaai2025} (\citeyear{stahlberg-et-al-aaai2025})
showed that the expressiveness of R-GNNs can go beyond $C_2$ simply by changing
the input encoding, and successfully learn general policies for domains that
require $C_3$ expressivity. However, these encodings are not practical for large
states as all possible object pairs need to be part of the input. Alternative
encodings and their impact on expressiveness have also been studied by
\citeauthor{horcik-sir-icaps2024} (\citeyear{horcik-sir-icaps2024}) and
\citeauthor{horcik-et-al-aaai2025} (\citeyear{horcik-et-al-aaai2025}).

Formally, the R-GNN input consists of a set of ground atoms $\mathcal{R}$
that describe the current state, the goal, and optionally one or more
successor states or actions. Different choices for $\mathcal{R}$ yield different encodings and different readout modalities. 
In our approach we adopt the current state $s$ encoding from \cite{stahlberg-et-al-kr2022}. Let
$\mathcal{O}$ be the set of objects in $s$, and let $\bar{o}$ denote a tuple of
objects. The atoms representing $s$ are given by:
\begin{align*}
    \mathcal{R}^{s} \;=\; &\{ P(\bar{o}) \mid P(\bar{o}) \in s \} \cup \{ P_{G,T}(\bar{o}) \mid P(\bar{o}) \in g \cap s \} 
    \cup \{ P_{G,F}(\bar{o}) \mid P(\bar{o}) \in g \setminus s \}.
\end{align*}
Goal predicates $P_{G,T}$ and $P_{G,F}$ are included to distinguish goal atoms
that are true in $s$ from those that are false. This distinction is crucial, as
R-GNNs may lack the expressiveness to infer these for predicates of arity greater than one
\citep{frances-et-al-aaai2021,chen-et-al-icaps2024,drexler-et-al-kr2024}.

\paragraph{External Encoding.}
This encoding serves as a baseline, first used by \cite{stahlberg-et-al-kr2023}
and later adapted for IW-policies by \cite{aichmueller-geffner-ijcai2025}. It
processes the current state $s$ and the successor state $s'$ separately,
combining them only at the readout stage to produce Q-values. Specifically, the
R-GNN generates embeddings for $\mathcal{R}^{s}$ and $\mathcal{R}^{s'}$
independently. Formally, let $\textrm{f}(o; \mathcal{R}^{s})$ and $\textrm{f}(o;
\mathcal{R}^{s'})$ be the embeddings of object $o$ after the final iteration
using $\mathcal{R}^{s}$ and $\mathcal{R}^{s'}$, respectively. The Q-value for a
transition to $s'$ is computed as:
\begin{equation*}
    Q(s, s') \;=\; \textrm{MLP}_1\left(\sum\nolimits_{o \in \mathcal{O}} \textrm{MLP}_2\left(\textrm{f}(o; \mathcal{R}^{s}), \textrm{f}(o; \mathcal{R}^{s'})\right)\right),
\end{equation*}
where $\textrm{MLP}_1$ and $\textrm{MLP}_2$ are multi-layer perceptrons.
Crucially, since the R-GNN processes states independently, it cannot compute
differences during message passing. The inner MLP combines the embeddings of
each object from both states, marking their first interaction.

\paragraph{Aggregated-Actions (AA) Encoding.}

This second baseline, presented by \citeauthor{stahlberg-geffner-aaai2026}
(\citeyear{stahlberg-geffner-aaai2026}), represents successor states via ground actions. Consequently, this
encoding applies only to learning flat policies that select primitive actions.
Formally, for a current state $s$, the input is:
\begin{equation*}
    \mathcal{R}^{\text{A}} \;=\; \mathcal{R}^{s} \cup \{ P_{A}(o_a, \bar{o}) \mid a \in \mathcal{A}[s], a = A(\bar{o}) \},
\end{equation*}
where $o_a$ is a new object representing ground action $a$. Here, $A$ is the
name of the action schema. The Q-value for ground action $a$ is
computed as:
\begin{equation*}
    Q(s, a) \;=\; \textrm{MLP}\left(\textrm{f}(o_a; \mathcal{R}^{\text{A}}), \sum\nolimits_{o \in \mathcal{O}} \textrm{f}(o; \mathcal{R}^{\text{A}})\right),
\end{equation*}
where $\textrm{f}(o_a; \mathcal{R}^{\text{A}})$ is the final embedding of the
action object $o_a$, and $\textrm{f}(o; \mathcal{R}^{\text{A}})$ are the final
embeddings of the original objects. Since the effects of ground action $a$ are
not explicitly represented, the R-GNN must infer them from the action name and
arguments.

\paragraph{Aggregated-Delta (AD) Encoding.}
 In our proposed encoding scheme of Aggregated-Delta, we move from representing each transition $s \to s'$ individually to instead representing the entire tree jointly. Let
$\mathcal{T} = (V, E)$ represent an IW (or its variants) directed lookahead tree from root state $s$. 

We represent the root state implicitly via the relation set $\mathcal{R}^s$ again. 
However, to compactly represent successors we introduce an explicit state object-node $o_{s'}$ for each $s' \in V \setminus \{s\}$ and encode only its added predicates $P^+(o_{s'}, \bar o)$ for atoms in $s' \setminus s$ and deleted predicates $P^-(o_{s'}, \bar o)$ for atoms in $s \setminus s'$. 
Crucially, the added and deleted predicates are anchored to their state object $o_{s'}$. 
This allows the R-GNN to compute embeddings for each state object $o_{s'}$ based on its differences from the root.

For each edge $(s', s'') \in E(\mathcal{T})$ a new atom $P_\mathrm{E}(o_{s'}, o_{s''})$ representing the transition is added that allows the successors to exchange information directly, instead of only indirectly via the problem's object nodes. Another benefit of encoding the tree as a whole is the option to represent relative search depth of successors in a structural way. To this end, we embed depth nodes $o_d$ for each depth $d \leq d_{\max}(\mathcal{T})$ and imply an order through auxiliary atoms $P_\mathrm{D}(o_d, o_{d'})$ whenever $d < d'$. Depth-information flows back into state-nodes via state-depth association atoms $P_{\mathrm{SD}}(o_{s'}, o_d)$ for each successor $s'$ at depth $d$. The complete set of atoms added to the input is then:
\begin{align*}
    \mathcal{R}^{\mathcal{T}} \;=\; &\mathcal{R}^{s} \cup \mathcal{R}^\mathcal{T}_E \cup \mathcal{R}_{D}^\mathcal{T} \cup \bigcup_{s' \in V(\mathcal{T}) \setminus \{s\}} \mathcal{R}^\mathcal{T}_{\Delta s'}, \text{ where } \\
    \mathcal{R}^\mathcal{T}_E \;=\; &\{ P_E(o_{s'}, o_{s''}) \mid (s', s'') \in E(\mathcal{T}) \}, \\
    \mathcal{R}_{D}^\mathcal{T} \;=\; &\{P_D(o_d, o_{d'}) \mid d < d'\} \cup \{P_{\mathrm{SD}}(o_{s'}, o_d) \mid s' \in V(\mathcal{T}) \textrm{ at depth } d\}, \\
    \mathcal{R}^\mathcal{T}_{\Delta s'} \;=\; &\{ P^+(o_{s'}, \bar{o}) \mid P(\bar{o}) \in s' \setminus s \} \; \cup \; \{ P^-(o_{s'}, \bar{o}) \mid P(\bar{o}) \in s \setminus s' \} \\
    \cup\; &\{ P^+_{G}(o_{s'}, \bar{o}) \mid P(\bar{o}) \in g \cap (s' \setminus s) \}
    \; \cup\; \{ P^-_{G}(o_{s'}, \bar{o}) \mid P(\bar{o}) \in g \cap (s \setminus s') \}.
\end{align*}
Finally, the Q-value to
state $s' \in V(\mathcal{T})$ is computed similarly as in AA:
\begin{equation*}
    Q(s, s') \;=\; \textrm{MLP}\left(\textrm{f}(o_{s'}; \mathcal{R}^{\mathcal{T}}), \sum\nolimits_{o \in \mathcal{O}} \textrm{f}(o; \mathcal{R}^{\mathcal{T}})\right).
\end{equation*}
Including both the state object embedding $\textrm{f}(o_{s'};
\mathcal{R}^{\mathcal{T}})$ and the aggregated original object embeddings
$\sum_{o \in \mathcal{O}} \textrm{f}(o; \mathcal{R}^{\mathcal{T}})$ is vital to
provide context about $s$ when evaluating the jump to $s'$. Although using only
the first term is possible, representing both states with a single embedding
proves difficult in practice. During readout, we compute Q-values for all states
$s' \in V(\mathcal{T})$ in a single batch.

AD improves resource usage in two ways. First, root-state embedding reuse reduces the number of required embeddings and the associated memory footprint by a factor of $|V(\mathcal{T})|$. Second, representing states through atom changes avoids message and gradient creation proportional to the number of atoms. As a result, IW-policy models that previously exceeded $24$GB of VRAM become trainable with at most $4$GB.

Between External and AD encodings, two meaningful interpolations are possible. The first, \emph{internal}, still encodes each transition $s \to s'$ separately, but represents the transition in a single graph over shared object nodes. It uses the full state description $\mathcal{R}^s_{\mathcal{P}}$ for $s$ and a duplicated predicate set $\mathcal{R}^{s'}_{\mathcal{P}'}$ for $s'$. The second, \emph{internal-delta}, also encodes each transition separately, but represents $s'$ by delta predicates $\mathcal{R}^s_{\Delta s'}$ without an explicit $o_{s'}$ state-node, while $s$ remains encoded through $\mathcal{R}^s$. Neither interpolation matches AD performance, since both lack tree-level communication between successor nodes. Still, both serve as ablations for the benefits of compact internalized inputs. Detailed results are provided in the appendix.
\section{Learning Generalized Policies}
\label{sec:learning}

The R-GNN is trained using deep Q-learning \citep{mnih-et-al-nature2015},
following the approach of \citeauthor{stahlberg-geffner-aaai2026}
(\citeyear{stahlberg-geffner-aaai2026}), which uses a lifted variant of hindsight relabeling.
The objective is to learn a Q-value function $Q(s, s')$ that estimates the expected cumulative reward of performing
a transition from state $s$ to successor state $s'$. In this paper, the reward
is always defined as $-1$. Thus, the Q-value approximately represents the
negative expected number of transitions to reach the goal from state $s$.

Additionally, AD encodings introduced depth information of successors as inductive bias which we enforce via an auxiliary loss. Such is beneficial whenever the distance of IW-jumps is crucial in an environment. For example, in the Spanner domain, the agent can only move forward and collect items on location, meaning bypassed items cannot be retrieved later. If the agent jumps to pick up the farthest spanner, it often guarantees a dead-end state. This loss ensures that distance information is properly learned, helping the agent acquire a strategy that avoids failure.
Since absolute label classification or value regression cannot generalize to values unseen during training, we enforce depth order by adapting the ranking loss of \citet{burges-et-al-icml2005} over $(o_k, o_l)$ pairs with $k < l$. To this end, a small probe matrix $W_\theta$ computes a score $z_{d} = W_\theta f({o_d})$ from each final depth-node embedding. Then, letting $w_{k,l} = 1 / (l -k)$ and $Z = \sum_{k < l \leq d_{\max}} w_{k,l}$ each such depth-node pair is compared in a rank-loss $\mathcal{L}_{\mathrm{d-rank}} =
    \frac{1}{Z}
    \sum_{k < l \leq d_{\max}}
    w_{k,l} \log\left(1 + \exp(z_{k} - z_{l})\right)$
where $d_{\max}$ is the maximum realized depth in that sample's IW tree. This enforces deeper nodes to receive larger scores relative to shallower ones in a generalizing manner.
Stochastic gradient descent updates the R-GNN parameters to minimize the aggregated loss over sampled mini-batches.

During training, exploration is encouraged using a Boltzmann policy, where the
probability of selecting a successor state $s'$ from state $s$ is given by $P(s' \mid s) \propto \exp(Q(s, s') / \tau)$,
where $\tau$ is the temperature parameter that controls the
exploration-exploitation trade-off.

\begin{table*}[t]
    \centering
    \footnotesize
    \setlength{\tabcolsep}{6pt}
    \renewcommand{\arraystretch}{0.73}
    \begin{tabular}{lrrrrrrrr}
\toprule
Domain & \textbf{AIW--AD} & IW--AD & AIW--Ext & IW--Ext & Flat--AA & LAMA* & Distincter* & GOOSE* \\
\midrule
blocksworld (90) & \textbf{\coverage{90}{100}} & \coverage{65}{72} & \coverage{78}{87} & \coverage{59}{66} & \coverage{76}{84} & \coverage{57}{63} & \coverage{88}{98} & \coverage{75}{83} \\
{\scriptsize \hspace*{1.25em}easy (30)} & {\scriptsize \textbf{\coverage{30}{100}}} & {\scriptsize \textbf{\coverage{30}{100}}} & {\scriptsize \textbf{\coverage{30}{100}}} & {\scriptsize \textbf{\coverage{30}{100}}} & {\scriptsize \textbf{\coverage{30}{100}}} & {\scriptsize \textbf{\coverage{30}{100}}} & {\scriptsize n/a} & {\scriptsize n/a} \\
{\scriptsize \hspace*{1.25em}medium (30)} & {\scriptsize \textbf{\coverage{30}{100}}} & {\scriptsize \textbf{\coverage{30}{100}}} & {\scriptsize \textbf{\coverage{30}{100}}} & {\scriptsize \coverage{29}{97}} & {\scriptsize \textbf{\coverage{30}{100}}} & {\scriptsize \coverage{24}{80}} & {\scriptsize n/a} & {\scriptsize n/a} \\
{\scriptsize \hspace*{1.25em}hard (30)} & {\scriptsize \textbf{\coverage{30}{100}}} & {\scriptsize \coverage{5}{17}} & {\scriptsize \coverage{18}{60}} & {\scriptsize \coverage{0}{0}} & {\scriptsize \coverage{16}{53}} & {\scriptsize \coverage{3}{10}} & {\scriptsize n/a} & {\scriptsize n/a} \\
\midrule
miconic (90) & \textbf{\coverage{90}{100}} & \textbf{\coverage{90}{100}} & \coverage{72}{80} & \coverage{72}{80} & \coverage{84}{93} & \textbf{\coverage{90}{100}} & \textbf{\coverage{90}{100}} & \textbf{\coverage{90}{100}} \\
{\scriptsize \hspace*{1.25em}easy (30)} & {\scriptsize \textbf{\coverage{30}{100}}} & {\scriptsize \textbf{\coverage{30}{100}}} & {\scriptsize \textbf{\coverage{30}{100}}} & {\scriptsize \textbf{\coverage{30}{100}}} & {\scriptsize \textbf{\coverage{30}{100}}} & {\scriptsize \textbf{\coverage{30}{100}}} & {\scriptsize n/a} & {\scriptsize n/a} \\
{\scriptsize \hspace*{1.25em}medium (30)} & {\scriptsize \textbf{\coverage{30}{100}}} & {\scriptsize \textbf{\coverage{30}{100}}} & {\scriptsize \textbf{\coverage{30}{100}}} & {\scriptsize \textbf{\coverage{30}{100}}} & {\scriptsize \textbf{\coverage{30}{100}}} & {\scriptsize \textbf{\coverage{30}{100}}} & {\scriptsize n/a} & {\scriptsize n/a} \\
{\scriptsize \hspace*{1.25em}hard (30)} & {\scriptsize \textbf{\coverage{30}{100}}} & {\scriptsize \textbf{\coverage{30}{100}}} & {\scriptsize \coverage{12}{40}} & {\scriptsize \coverage{12}{40}} & {\scriptsize \coverage{24}{80}} & {\scriptsize \textbf{\coverage{30}{100}}} & {\scriptsize n/a} & {\scriptsize n/a} \\
\midrule
spanner (90) & \textbf{\coverage{90}{100}} & \coverage{61}{68} & \coverage{32}{36} & \coverage{55}{61} & \textbf{\coverage{90}{100}} & \coverage{30}{33} & \textbf{\coverage{90}{100}} & \coverage{73}{81} \\
{\scriptsize \hspace*{1.25em}easy (30)} & {\scriptsize \textbf{\coverage{30}{100}}} & {\scriptsize \textbf{\coverage{30}{100}}} & {\scriptsize \textbf{\coverage{30}{100}}} & {\scriptsize \coverage{27}{90}} & {\scriptsize \textbf{\coverage{30}{100}}} & {\scriptsize \textbf{\coverage{30}{100}}} & {\scriptsize n/a} & {\scriptsize n/a} \\
{\scriptsize \hspace*{1.25em}medium (30)} & {\scriptsize \textbf{\coverage{30}{100}}} & {\scriptsize \textbf{\coverage{30}{100}}} & {\scriptsize \coverage{2}{7}} & {\scriptsize \coverage{27}{90}} & {\scriptsize \textbf{\coverage{30}{100}}} & {\scriptsize \coverage{0}{0}} & {\scriptsize n/a} & {\scriptsize n/a} \\
{\scriptsize \hspace*{1.25em}hard (30)} & {\scriptsize \textbf{\coverage{30}{100}}} & {\scriptsize \coverage{1}{3}} & {\scriptsize \coverage{0}{0}} & {\scriptsize \coverage{1}{3}} & {\scriptsize \textbf{\coverage{30}{100}}} & {\scriptsize \coverage{0}{0}} & {\scriptsize n/a} & {\scriptsize n/a} \\
\midrule
satellite (90) & \coverage{88}{98} & \coverage{55}{61} & \coverage{61}{68} & \coverage{56}{62} & \coverage{51}{57} & \textbf{\coverage{90}{100}} & \coverage{48}{53} & \coverage{53}{59} \\
{\scriptsize \hspace*{1.25em}easy (30)} & {\scriptsize \textbf{\coverage{30}{100}}} & {\scriptsize \textbf{\coverage{30}{100}}} & {\scriptsize \textbf{\coverage{30}{100}}} & {\scriptsize \textbf{\coverage{30}{100}}} & {\scriptsize \textbf{\coverage{30}{100}}} & {\scriptsize \textbf{\coverage{30}{100}}} & {\scriptsize n/a} & {\scriptsize n/a} \\
{\scriptsize \hspace*{1.25em}medium (30)} & {\scriptsize \textbf{\coverage{30}{100}}} & {\scriptsize \coverage{25}{83}} & {\scriptsize \textbf{\coverage{30}{100}}} & {\scriptsize \coverage{26}{87}} & {\scriptsize \coverage{21}{70}} & {\scriptsize \textbf{\coverage{30}{100}}} & {\scriptsize n/a} & {\scriptsize n/a} \\
{\scriptsize \hspace*{1.25em}hard (30)} & {\scriptsize \coverage{28}{93}} & {\scriptsize \coverage{0}{0}} & {\scriptsize \coverage{1}{3}} & {\scriptsize \coverage{0}{0}} & {\scriptsize \coverage{0}{0}} & {\scriptsize \textbf{\coverage{30}{100}}} & {\scriptsize n/a} & {\scriptsize n/a} \\
\midrule
transport (90) & \textbf{\coverage{86}{96}} & \coverage{74}{82} & \coverage{55}{61} & \coverage{66}{73} & \coverage{74}{82} & \coverage{72}{80} & \coverage{50}{56} & \coverage{29}{32} \\
{\scriptsize \hspace*{1.25em}easy (30)} & {\scriptsize \textbf{\coverage{30}{100}}} & {\scriptsize \textbf{\coverage{30}{100}}} & {\scriptsize \coverage{29}{97}} & {\scriptsize \textbf{\coverage{30}{100}}} & {\scriptsize \coverage{29}{97}} & {\scriptsize \textbf{\coverage{30}{100}}} & {\scriptsize n/a} & {\scriptsize n/a} \\
{\scriptsize \hspace*{1.25em}medium (30)} & {\scriptsize \coverage{29}{97}} & {\scriptsize \textbf{\coverage{30}{100}}} & {\scriptsize \coverage{19}{63}} & {\scriptsize \textbf{\coverage{30}{100}}} & {\scriptsize \coverage{25}{83}} & {\scriptsize \textbf{\coverage{30}{100}}} & {\scriptsize n/a} & {\scriptsize n/a} \\
{\scriptsize \hspace*{1.25em}hard (30)} & {\scriptsize \textbf{\coverage{27}{90}}} & {\scriptsize \coverage{14}{47}} & {\scriptsize \coverage{7}{23}} & {\scriptsize \coverage{6}{20}} & {\scriptsize \coverage{20}{67}} & {\scriptsize \coverage{12}{40}} & {\scriptsize n/a} & {\scriptsize n/a} \\
\midrule
ferry (90) & \coverage{80}{89} & \coverage{76}{84} & \coverage{71}{79} & \coverage{67}{74} & \coverage{62}{69} & \coverage{74}{82} & \textbf{\coverage{83}{92}} & \coverage{76}{84} \\
{\scriptsize \hspace*{1.25em}easy (30)} & {\scriptsize \textbf{\coverage{30}{100}}} & {\scriptsize \textbf{\coverage{30}{100}}} & {\scriptsize \textbf{\coverage{30}{100}}} & {\scriptsize \textbf{\coverage{30}{100}}} & {\scriptsize \coverage{29}{97}} & {\scriptsize \textbf{\coverage{30}{100}}} & {\scriptsize n/a} & {\scriptsize n/a} \\
{\scriptsize \hspace*{1.25em}medium (30)} & {\scriptsize \textbf{\coverage{30}{100}}} & {\scriptsize \textbf{\coverage{30}{100}}} & {\scriptsize \textbf{\coverage{30}{100}}} & {\scriptsize \textbf{\coverage{30}{100}}} & {\scriptsize \coverage{29}{97}} & {\scriptsize \textbf{\coverage{30}{100}}} & {\scriptsize n/a} & {\scriptsize n/a} \\
{\scriptsize \hspace*{1.25em}hard (30)} & {\scriptsize \textbf{\coverage{20}{67}}} & {\scriptsize \coverage{16}{53}} & {\scriptsize \coverage{11}{37}} & {\scriptsize \coverage{7}{23}} & {\scriptsize \coverage{4}{13}} & {\scriptsize \coverage{14}{47}} & {\scriptsize n/a} & {\scriptsize n/a} \\
\midrule
rovers (90) & \coverage{67}{74} & \coverage{62}{69} & \coverage{24}{27} & \coverage{36}{40} & \coverage{27}{30} & \textbf{\coverage{78}{87}} & \coverage{42}{47} & \coverage{37}{41} \\
{\scriptsize \hspace*{1.25em}easy (30)} & {\scriptsize \textbf{\coverage{30}{100}}} & {\scriptsize \textbf{\coverage{30}{100}}} & {\scriptsize \coverage{24}{80}} & {\scriptsize \textbf{\coverage{30}{100}}} & {\scriptsize \coverage{27}{90}} & {\scriptsize \textbf{\coverage{30}{100}}} & {\scriptsize n/a} & {\scriptsize n/a} \\
{\scriptsize \hspace*{1.25em}medium (30)} & {\scriptsize \coverage{24}{80}} & {\scriptsize \coverage{25}{83}} & {\scriptsize \coverage{0}{0}} & {\scriptsize \coverage{6}{20}} & {\scriptsize \coverage{0}{0}} & {\scriptsize \textbf{\coverage{30}{100}}} & {\scriptsize n/a} & {\scriptsize n/a} \\
{\scriptsize \hspace*{1.25em}hard (30)} & {\scriptsize \coverage{13}{43}} & {\scriptsize \coverage{7}{23}} & {\scriptsize \coverage{0}{0}} & {\scriptsize \coverage{0}{0}} & {\scriptsize \coverage{0}{0}} & {\scriptsize \textbf{\coverage{18}{60}}} & {\scriptsize n/a} & {\scriptsize n/a} \\
\midrule
childsnack (90) & \textbf{\coverage{64}{71}} & \coverage{33}{37} & \coverage{47}{52} & \coverage{37}{41} & \coverage{41}{46} & \coverage{36}{40} & \textbf{\coverage{64}{71}} & \coverage{29}{32} \\
{\scriptsize \hspace*{1.25em}easy (30)} & {\scriptsize \textbf{\coverage{30}{100}}} & {\scriptsize \coverage{28}{93}} & {\scriptsize \textbf{\coverage{30}{100}}} & {\scriptsize \textbf{\coverage{30}{100}}} & {\scriptsize \coverage{27}{90}} & {\scriptsize \textbf{\coverage{30}{100}}} & {\scriptsize n/a} & {\scriptsize n/a} \\
{\scriptsize \hspace*{1.25em}medium (30)} & {\scriptsize \textbf{\coverage{30}{100}}} & {\scriptsize \coverage{5}{17}} & {\scriptsize \coverage{17}{57}} & {\scriptsize \coverage{7}{23}} & {\scriptsize \coverage{12}{40}} & {\scriptsize \coverage{5}{17}} & {\scriptsize n/a} & {\scriptsize n/a} \\
{\scriptsize \hspace*{1.25em}hard (30)} & {\scriptsize \textbf{\coverage{4}{13}}} & {\scriptsize \coverage{0}{0}} & {\scriptsize \coverage{0}{0}} & {\scriptsize \coverage{0}{0}} & {\scriptsize \coverage{2}{7}} & {\scriptsize \coverage{1}{3}} & {\scriptsize n/a} & {\scriptsize n/a} \\
\midrule
sokoban (90) & \coverage{13}{14} & \coverage{13}{14} & \coverage{17}{19} & \coverage{15}{17} & \coverage{14}{16} & \textbf{\coverage{44}{49}} & \coverage{32}{36} & \coverage{38}{42} \\
{\scriptsize \hspace*{1.25em}easy (30)} & {\scriptsize \coverage{13}{43}} & {\scriptsize \coverage{13}{43}} & {\scriptsize \coverage{17}{57}} & {\scriptsize \coverage{15}{50}} & {\scriptsize \coverage{14}{47}} & {\scriptsize \textbf{\coverage{30}{100}}} & {\scriptsize n/a} & {\scriptsize n/a} \\
{\scriptsize \hspace*{1.25em}medium (30)} & {\scriptsize \coverage{0}{0}} & {\scriptsize \coverage{0}{0}} & {\scriptsize \coverage{0}{0}} & {\scriptsize \coverage{0}{0}} & {\scriptsize \coverage{0}{0}} & {\scriptsize \textbf{\coverage{14}{47}}} & {\scriptsize n/a} & {\scriptsize n/a} \\
{\scriptsize \hspace*{1.25em}hard (30)} & {\scriptsize \coverage{0}{0}} & {\scriptsize \coverage{0}{0}} & {\scriptsize \coverage{0}{0}} & {\scriptsize \coverage{0}{0}} & {\scriptsize \coverage{0}{0}} & {\scriptsize \coverage{0}{0}} & {\scriptsize n/a} & {\scriptsize n/a} \\
\midrule
floortile (90) & \coverage{0}{0} & \coverage{0}{0} & \coverage{0}{0} & \coverage{0}{0} & \coverage{0}{0} & \textbf{\coverage{15}{17}} & \coverage{2}{2} & \coverage{2}{2} \\
{\scriptsize \hspace*{1.25em}easy (30)} & {\scriptsize \coverage{0}{0}} & {\scriptsize \coverage{0}{0}} & {\scriptsize \coverage{0}{0}} & {\scriptsize \coverage{0}{0}} & {\scriptsize \coverage{0}{0}} & {\scriptsize \textbf{\coverage{15}{50}}} & {\scriptsize n/a} & {\scriptsize n/a} \\
{\scriptsize \hspace*{1.25em}medium (30)} & {\scriptsize \coverage{0}{0}} & {\scriptsize \coverage{0}{0}} & {\scriptsize \coverage{0}{0}} & {\scriptsize \coverage{0}{0}} & {\scriptsize \coverage{0}{0}} & {\scriptsize \coverage{0}{0}} & {\scriptsize n/a} & {\scriptsize n/a} \\
{\scriptsize \hspace*{1.25em}hard (30)} & {\scriptsize \coverage{0}{0}} & {\scriptsize \coverage{0}{0}} & {\scriptsize \coverage{0}{0}} & {\scriptsize \coverage{0}{0}} & {\scriptsize \coverage{0}{0}} & {\scriptsize \coverage{0}{0}} & {\scriptsize n/a} & {\scriptsize n/a} \\
\midrule
\noalign{\vskip-\belowrulesep}
\rowcolor{black!8!white}
\btcoloredrowstrut\emph{Total (900)}
& \textbf{\coverage{668}{74}} & \coverage{529}{59} & \coverage{457}{51} & \coverage{463}{51} & \coverage{519}{58} & \coverage{586}{65} & \coverage{589}{65} & \coverage{502}{56} \\
\noalign{\vskip-\aboverulesep}
\bottomrule
\end{tabular}

    \caption{%
        IPC 2023 benchmark coverage results for AIW with AD encoding (highlighted in bold) against ablations and baselines. Methods marked * perform exponential search at test time. Numbers in parentheses are percentages of row totals.}%
    \label{tab:coverage:all}
\end{table*}

\section{Experiments}
\label{sec:experiments}

We detail the experimental benchmarks, training and testing setup, and discuss the results.

\paragraph{Domains and Data.}

Before diving into the specifics, it is important to contextualize these benchmarks, as our evaluation differs from standard practices in deep learning and RL. First, our objective is strictly binary goal-reaching from an initial state in a goal-conditioned RL setting with extremely sparse rewards; unlike typical RL setups, there is no accumulated reward signal or partial degree of success. Second, we focus on zero-shot structural generalization: models are trained on small instances and tested on significantly larger ones, often featuring more than 10--20 times the
number of objects seen during training and an exponentially larger state space. For example, in the Blocksworld domain, training is performed on instances with up to 20 blocks and testing on instances with up to 488 blocks. In contrast, standard ML benchmarks typically evaluate on test distributions of similar scale to the training set. Finally, to rigorously test the robustness of the learned representations, we evaluate using strictly greedy, deterministic policies. While we use a limited lookahead search, the agent must commit to its ``jumps'' without the safety net of backtracking or multiple stochastic rollouts common in other RL approaches. This ensures we are measuring the true generalization capacity of the learned policy itself, rather than the compensatory power of an extensive search algorithm.

We evaluate our approach on two benchmark sets: hard-to-learn domains from
\citet{drexler-et-al-kr2024} that require
more than $C_2$ logic expressivity, and the IPC Learning Track 2023
\citep{taitler-et-al-aimag2024}. This means that both
training and testing has been carried out in instances that are much larger than
those used in previous works on general policy learning. Detailed domain descriptions and an overview of the train, validation
and test sets are available in the appendix, where we also provide a visualization of the immense scaling factors present in the IPC benchmark.

\paragraph{Training.}
Our implementation is based on PyTorch and extends the codebase of
\citet{stahlberg-geffner-aaai2026}.
The architecture remains unchanged; methods differ only in their input encoding
and hyperparameters.
We set the embedding size to $32$ and used smooth maximum aggregation. The
learning rate decayed linearly from $10^{-3}$ to $10^{-5}$ over $300$ episodes.
Similarly, the Boltzmann temperature decayed linearly from $1.0$ to $0.1$ over
$1000$ episodes. The discount factor was set to $0.999$. Each episode consisted
of $32$ optimization steps with a batch size of $32$. The replay buffer size was
$100$ for IW-policies and $1000$ for flat-policies. We generated $4$ trajectories per episode, each capped at $20$ IW jumps or $200$ action steps. We
trained $5$ seeds per method and domain on an cluster with an Intel Xeon
Platinum 8352M CPU, $20$~GB RAM, and an NVIDIA A10/L40S GPU capped at $24$GB, with a time limit of $12$ hours. Model selection was based on validation coverage; in case of ties,
we prioritized the shortest total solution length, followed by the lowest total
TD-error, and finally random selection. We test the best $5$ checkpoints among all runs and report the best outcome.

\paragraph{Testing.}

We evaluate policies on the test set using a greedy strategy over
Q-value functions. We consider two types of goal-conditioned policies: flat policies that select primitive actions, and IW-policies that select IW-jumps as described in the background. At each step, the policy selects the successor
state $s^* = \arg\max_{s' \in \mathcal{S}} Q(s, s')$ that maximizes the Q-value
among all valid successors $\mathcal{S}$. To ensure progress and avoid infinite
loops, we maintain a history of visited states within the current episode and
exclude any transition to a previously visited state from the
candidate successors. The evaluation continues until the goal condition is
satisfied ($g \subseteq s$), or until a resource limit is exhausted. We enforce a limit of $1,000$ choices and a timeout of $15$ and $60$ minutes per instance for flat- and IW-policies, respectively. 
We test the following methods:
\begin{enumerate}[(a)]
    \item \emph{AIW--AD}: Our full proposed method of combining AIW with AD encoding.
    \item \emph{IW--AD}: IW-policies \citep{aichmueller-geffner-ijcai2025} using our proposed AD encoding. This method compares the unidimensional improvement made by more efficient encodings.
    \item \emph{AIW Ext}: AIW-policies using the baseline's External encoding. This method compares the unidimensional improvement made by the relaxed IW-lookahead variant.
    \item \emph{IW Ext}: IW-policies using External encodings, the direct baseline.
    \item \emph{Flat AA}: This baseline uses the AA encoding \citep{stahlberg-geffner-aaai2026} (see Section~\ref{sec:encodings}) and relies entirely on the greedy selection of primitive actions that to not result in a cycle.
    \item \emph{LAMA} -- The classical planner LAMA \citep{richter-westphal-ipc2008} which uses a best-first search to first find a plan and then improves it using iterated weighted $A^\star$ search. LAMA was given 1 hour and 90 GB of memory per instance.
    \item \emph{Distincter}: Distincter \citet{bai-et-al-icaps2025} uses Greedy Best-First Search (GBFS) with GNNs that provides both heuristic guidance and approximate symmetry detection. To prioritize speed, the GNNs rely on a shallow 3–4 layer architecture; however, this restricted expressivity risks falsely identifying distinct states or actions as symmetric.
    \item  \emph{GOOSE}: WL-GOOSE \citet{chen-et-al-icaps2024} diverges from deep learning approaches by using the WL algorithm alongside classical machine learning models to learn search heuristics from scratch. Rather than relying on GNNs, it colors a graph (specifically, the ILG) and uses the resulting color counts to define its heuristic. This strikes a good balance between computational efficiency and heuristic quality, resulting in competitive performance when used with GBFS.
\end{enumerate}
We also tested BAIW, C-AIW, and Flat policies using AD encoding in the appendix.
We note that all (A)IW and flat policy methods -- baseline or not -- are trained via the described learning method in this paper, including the usage of lifted HER \citep{stahlberg-geffner-aaai2026}.
Technically, extending the IW(1) policies of \cite{aichmueller-geffner-ijcai2025} with lifted HER is also first presented in this paper, yet we consider this straightforward.
The results for Distincter and WL-GOOSE -- trained with supervised learning -- are taken directly from their original papers, which used a timeout of 30 minutes and 8 GB of memory.

\paragraph{Results}

Table~\ref{tab:coverage:all} reports coverage on the IPC 2023 benchmark.
In summary, AIW--AD yields the strongest overall coverage among all methods and improves substantially over every baseline.
It also surpasses the classical planning baseline LAMA by a wide margin.
To our knowledge, the only previous method to exceed LAMA on this benchmark is \cite{bai-et-al-icaps2025}, and only by a small margin.
Here, the gain over LAMA is obtained \emph{without resorting to an exponential search}: AIW runs in time linear in the number of objects and is invoked only a bounded number of times in sequence.

The main remaining failure cases are \emph{Sokoban} and \emph{Floortile}.
Sokoban is PSPACE-complete \citep{culberson-tr1997}, making it difficult to learn a Q-function that generalizes well enough to replace search.
This difficulty is reflected in the results, where only search-based methods solve a non-trivial subset of its instances.
Floortile poses a different challenge: under our Q-learning setup with Hindsight Experience Replay (HER), the domain provides a weak training signal.
Success requires precise action sequences to avoid dead ends, but failed trajectories are rarely relabeled into useful supervision for learning such sequences.
However, no evaluated method performs convincingly on either domain.

Outside these puzzle domains, performance is consistently strong.
In nearly all other domains, AIW--AD solves all easy and medium instances, and most hard instances, with the main remaining gaps concentrated in Rovers and Childsnack.
Childsnack exposes a scaling regime that is prohibitive for methods that choose discriminatively from an explicitly enumerated set of options.
On the hardest instances, the branching factor reaches tens of millions of applicable actions. Mere action enumeration requires up to 10 minutes, despite highly optimized software and latest hardware, rendering decision-making from fully enumerated options unlikely to scale under reasonable budgets.
C-AIW solves six additional hard instances and raises coverage in this domain to $70/90$, the best result to date, to the best of our knowledge.
Yet, further progress on Childsnack will likely require models that opt to generate decisions compositionally rather than select them from an enumerated set.

These results show that abstracted width-based lookaheads improve scalability, while also preserving a useful subgoal structure that learned models can exploit.
Previous work showed that IW lookaheads can simplify the cost structure and lower the expressive requirements for learned policies \citep{aichmueller-geffner-ijcai2025}.
Our results show that this effect persists at scale and also under abstracted width-1 lookaheads, despite more aggressive pruning.
This is most evident in Satellite and Rovers, two domains known to require features beyond the $C_2$ logic fragment \citep{drexler-et-al-kr2024,stahlberg-et-al-icaps2022}.
Satellite is fully solved by AIW--AD. On Rovers, some hard instances remain unresolved, but the gains over previous learned approaches are substantial.
Moreover, performance improves steadily as the local scope is expanded from primitive successors to BAIW, C-AIW, and finally AIW successors, underscoring the value of principled lookaheads.

The same pattern carries over to a list of domains not included in the IPC 2023 benchmark (see Table \ref{tab:coverage:custom-main}, appendix), especially containing Grid and Logistics, which also lie beyond $C_2$ logic.
There, more aggressive abstraction failed: BAIW reaches only $48$\% test coverage in Logistics due to pruning too aggressively. However, it still achieves full coverage on Grid.
By contrast, AIW--AD attains full coverage on both domains again. Together, these results show that AIW--AD is not only effective on the IPC 2023 benchmark, but also on a vast set of structurally diverse domains.

\section{Conclusions}
\label{sec:conclusions}

We introduced two main contributions.
First, we proposed Aggregated-Delta (AD) encodings for R-GNNs, which represent transitions through state differences and combine all lookahead successors into a single relational graph.
This enables IW-based successors to be evaluated in one forward pass, greatly reducing memory and runtime costs and removing the main scalability bottleneck of previous IW-policy formulations.
Second, we introduced Abstracted IW (AIW), a stricter IW variant that reduces the novelty-based search space by modifying novelty calculation itself.

Together, these contributions yield a new state of the art in generalized planning.
On the IPC 2023 benchmark, AIW combined with AD significantly outperforms all previous baselines, including the strong baseline LAMA.
The key point is not only that the learned policy improves, but also that AD makes larger lookaheads practical, especially with AIW, whose novelty search scales with object counts rather than atom counts.
At the same time, the results show that the stronger abstractions used by AIW still preserve meaningful subgoals that learned models can exploit.
The gains on Satellite, Rovers, Grid, and Logistics -- domains that require beyond-$C_2$ expressiveness -- suggest that AIW reduces the expressive burden on the architecture in two ways.
First, AIW induces a simpler cost structure by replacing sequences of primitive decisions with jump endpoints.
Second, because intermediate states inside such jumps are handled by the lookahead rather than by the network, the learned policy can avoid distinctions that would otherwise require features beyond $C_2$.
Together, these effects show that efficient width-based lookahead enable general policies to scale not only by reducing search cost, but also by reducing what the learned architecture must represent.

\clearpage

\begin{ack}
We thank Jonas Gösgens and Niklas Jansen for their helpful discussions and insights during the development of this work.
The research has been supported by the Alexander von Humboldt Foundation with funds from the Federal Ministry for Education and Research, Germany.
This project has received funding from the European Research Council (ERC) under the European Union's Horizon 2020 research and innovations programme (Grant agreement No. 885107).
This project was also funded by the German Federal Ministry of Education and Research (BMBF) and the Ministry of Culture and Science of the German State of North Rhine-Westphalia (MKW) under the Excellence Strategy of the Federal Government and the L\"ander.
\end{ack}
\bibliographystyle{plainnat}
\bibliography{bibliography}

\clearpage
\appendix

\section{Benchmark Results}
We complement in table \ref{tab:coverage:custom-main} the reporting of test outcomes on our benchmark for methods described in the experimental section. 
\begin{table*}[t]
    \centering
    \footnotesize
    \renewcommand{\arraystretch}{0.73}
    \begin{tabular}{lrrrrrr}
\toprule
Domain & AIW--AD & IW--AD & AIW--Ext & IW--Ext & Flat AA & LAMA \\
\midrule
depots (86) & \textbf{\coverage{86}{100}} & \textbf{\coverage{86}{100}} & \coverage{83}{97} & \coverage{81}{94} & \coverage{76}{88} & \coverage{85}{99} \\
\midrule
delivery (80) & \textbf{\coverage{80}{100}} & \textbf{\coverage{80}{100}} & \textbf{\coverage{80}{100}} & \textbf{\coverage{80}{100}} & \coverage{25}{31} & \coverage{79}{99} \\
\midrule
driverlog (123) & \coverage{120}{98} & \textbf{\coverage{123}{100}} & \coverage{116}{94} & \textbf{\coverage{123}{100}} & \coverage{122}{99} & \textbf{\coverage{123}{100}} \\
\midrule
goldminer (359) & \textbf{\coverage{359}{100}} & \textbf{\coverage{359}{100}} & \coverage{133}{37} & \coverage{69}{19} & \coverage{132}{37} & \coverage{222}{62} \\
\midrule
grippers (312) & \textbf{\coverage{312}{100}} & \textbf{\coverage{312}{100}} & \textbf{\coverage{312}{100}} & \textbf{\coverage{312}{100}} & \textbf{\coverage{312}{100}} & \textbf{\coverage{312}{100}} \\
\midrule
freecell (117) & \textbf{\coverage{117}{100}} & \textbf{\coverage{117}{100}} & \textbf{\coverage{117}{100}} & \textbf{\coverage{117}{100}} & \textbf{\coverage{117}{100}} & \textbf{\coverage{117}{100}} \\
\midrule

visitall (100) & \textbf{\coverage{100}{100}} & \textbf{\coverage{100}{100}} & \textbf{\coverage{100}{100}} & \textbf{\coverage{100}{100}} & \coverage{97}{97} & \textbf{\coverage{100}{100}} \\
\midrule
reward (99) & \textbf{\coverage{99}{100}} & \textbf{\coverage{99}{100}} & \textbf{\coverage{99}{100}} & \textbf{\coverage{99}{100}} & \coverage{97}{98} & \textbf{\coverage{99}{100}} \\
\midrule
grid* (100) & \textbf{\coverage{100}{100}} & \coverage{99}{99} & \textbf{\coverage{100}{100}} & \coverage{96}{96} & \coverage{27}{27} & \textbf{\coverage{100}{100}} \\
\midrule
logistics* (100) & \coverage{97}{97} & \coverage{98}{98} & \coverage{0}{0} & \coverage{7}{7} & \coverage{25}{25} & \textbf{\coverage{100}{100}} \\
\midrule
satellite* (100) & \textbf{\coverage{100}{100}} & \textbf{\coverage{100}{100}} & \coverage{26}{26} & \coverage{29}{29} & \coverage{73}{73} & \textbf{\coverage{100}{100}} \\
\midrule
Total (1576) & \textbf{\coverage{1570}{100}} & \coverage{1573}{100} & \coverage{1166}{74} & \coverage{1113}{71} & \coverage{1103}{70} & \coverage{1437}{91} \\
\bottomrule
\end{tabular}

    \caption{%
        Benchmark coverage results for AIW--AD against its baselines and unidimensional improvement variants. Domains marked * are known to require logic beyond $C_2$ expressiveness. Numbers in parentheses are percentages of row totals.}%
    \label{tab:coverage:custom-main}
\end{table*}
We can clearly read two observations off the table: (1) AD encodings enhance the R-GNN's expressivity, and (2) width-based looakaheads simplify problem structures. 

AIW--AD achieves full coverage across all domains, particularly even for those requiring logic beyond $C_2$. The same holds only for IW--AD showing that AD input encodings directly enhance the model's expressivity. We can observe these facts hold for both IW and AIW in our benchmark distinctly, since its test-time scaling is more moderate and, thus, does not render IW(1) itself inefficient, unlike the IPC 2023 benchmark. On one domain, Goldminer, only AD encoded lookaheads are able to produce solutions for the entire set, again surpassing even LAMA. AA-encoded and AD-encoded (see table \ref{tab:coverage:custom-ablation}) flat policies are not able to achieve sufficient solve rates on many domains, reaffirming the observation made by \citet{aichmueller-geffner-ijcai2025} that (A)IW lookaheads simplify problem structure indeed and that this simplification is crucial for solving some domains.

\section{Base-Abstracted IW and Capacity-Limited Abstracted IW}

As mentioned in section \ref{sec:iw}, we also considered two modifications of AIW. A fallback variant of AIW in Base-Abstracted IW, which is the backup abstraction whenever a domain exhibits no typing information, and a depth-capacity limiting form in C-AIW. The following will further contextualize the methods and then discuss their results on the ICP 2023 benchmark as found in table \ref{tab:coverage:riw-variants} alongside Flat AD.
\paragraph{Base-Abstracted IW (BAIW).}
Reducing every other object information in an atom $P(o_1, \dots, o_n)$ to the base \texttt{\emph{object}}-type erases the object's information. Alternatively, BAIW can be seen as a novelty check via iteratively testing each of the $n$ position-derived unary atoms $P_i(o_i)$.
As a lookahead BAIW is also generally effective, but can be too coarse when several semantically different object types occur in the same predicate. The Spanner domain provides a simple example for this. The first argument $x$ of its predicate $$\texttt{at(x - \texttt{\emph{locatable}}, l - \texttt{\emph{location}})}$$ applies both to the \texttt{\emph{man}}-typed object \texttt{bob} and to spanner objects, since both are of a subtype of \texttt{\emph{locatable}}. Under the base abstraction, an atom \texttt{at(bob, loc)} is tested through the abstracted atoms \texttt{at(bob, *)} and \texttt{at(*, loc)}, where \texttt{*} denotes the object compressed to the base \texttt{\emph{object}}-type. The first abstracted atom, \texttt{at(bob, *)}, is already seen in the root through \texttt{at(bob, loc0)}. The second abstracted atom, \texttt{at(*, loc)}, can also already be seen in the root, because spanners occupy locations and generate atoms such as \texttt{at(spanner1, loc)}. Consequently, moving \texttt{bob} to a new location may fail to produce any novel atom. The AIW lookahead would be defunct, \ie, not provide any expansion of the root, if not for the primitive successors guaranteed by a prior IW(0) search, as the logic of IW(1) implies. A circumvention of this is to allow the primitive successors to be marked as novel themselves, to ensure BAIW continues novelty checks from these states, effectively combining IW(0) and BAIW(1) lookaheads in one go. This extension also provided meaningful results in preliminary experiments, but we did not evaluate it further.

AIW prevents this failure mode naturally by retaining the type of compressed objects. Then, \texttt{at(bob, loc)} is tested through the abstracted atoms \texttt{at(bob, \emph{location})} and \texttt{at(\emph{man}, loc)}. A location previously seen only through \texttt{at(spanner1, loc)} contributes \texttt{at(\texttt{\emph{locatable}}, loc)}, but not the more specific \texttt{at(\texttt{\emph{man}}, loc)}. Hence, \texttt{at(\texttt{\emph{man}}, loc)} remains novel when \texttt{bob} first reaches that location. Retaining types in the abstraction thus preserves enough role information to prevent the lookahead from collapsing, while staying meaningful, but coarser than full IW(1).

\paragraph{Capacity-Limited-AIW (C-AIW).}
To further account for infeasibly large branching factors, we also tried coupling the search with a layer capacity limit $C$, so that at any depth $d$ in the BFS we may at most retain $C$ novel successors to expand at depth $d+1$. Selection of the $C$ candidates happens via a scoring heuristic $h(s')$ that counts the number of satisfied goals in $s'$ and prefers those of higher score, breaking ties on a first-seen basis. We experimented with various values for $C$, finding $C=1000$ to work well in experiments.
Although Capacity-Limited AIW (C-AIW) further weakens completeness for width-1 goals, we gain an effective search limiter that reduces successor counts based on a fast and simple heuristic.
 In general, capacity limitation can lose crucial successors as shown in table \ref{tab:coverage:riw-variants} with Satellite, Rovers, and Transport, in which performance drops for 'hard' problems, \ie, the largest ones, where capacity limits actually start ejecting successors. In turn, capacity limits can improve the search when there are many equivalent continuation paths to the goal, \eg, due to action commutativity or argument permutations, as is common in Spanner and Childsnack. Its result of 70/90 solved instances on Childsnack is the highest reported rate on this domain to date. 

\begin{table*}[t]
    \centering
    \footnotesize
    \renewcommand{\arraystretch}{0.73}
    \begin{tabular}{lrrrrr}
\toprule
Domain & AIW & BAIW & C-AIW & Flat AD & Flat AA \\
\midrule
blocksworld (90) & \coverage{90}{100} & \coverage{89}{99} & \textbf{\coverage{90}{100}} & \coverage{76}{84} & \coverage{76}{84} \\
{\scriptsize \hspace*{1.25em}easy (30)} & {\scriptsize \textbf{\coverage{30}{100}}} & {\scriptsize \textbf{\coverage{30}{100}}} & {\scriptsize \textbf{\coverage{30}{100}}} & {\scriptsize \textbf{\coverage{30}{100}}} & {\scriptsize \textbf{\coverage{30}{100}}} \\
{\scriptsize \hspace*{1.25em}medium (30)} & {\scriptsize \textbf{\coverage{30}{100}}} & {\scriptsize \textbf{\coverage{30}{100}}} & {\scriptsize \textbf{\coverage{30}{100}}} & {\scriptsize \textbf{\coverage{30}{100}}} & {\scriptsize \textbf{\coverage{30}{100}}} \\
{\scriptsize \hspace*{1.25em}hard (30)} & {\scriptsize \coverage{30}{100}} & {\scriptsize \coverage{29}{97}} & {\scriptsize \textbf{\coverage{30}{100}}} & {\scriptsize \coverage{16}{53}} & {\scriptsize \coverage{16}{53}} \\
\midrule
miconic (90) & \textbf{\coverage{90}{100}} & \textbf{\coverage{90}{100}} & \textbf{\coverage{90}{100}} & \coverage{84}{93} & \coverage{84}{93} \\
{\scriptsize \hspace*{1.25em}easy (30)} & {\scriptsize \textbf{\coverage{30}{100}}} & {\scriptsize \textbf{\coverage{30}{100}}} & {\scriptsize \textbf{\coverage{30}{100}}} & {\scriptsize \textbf{\coverage{30}{100}}} & {\scriptsize \textbf{\coverage{30}{100}}} \\
{\scriptsize \hspace*{1.25em}medium (30)} & {\scriptsize \textbf{\coverage{30}{100}}} & {\scriptsize \textbf{\coverage{30}{100}}} & {\scriptsize \textbf{\coverage{30}{100}}} & {\scriptsize \textbf{\coverage{30}{100}}} & {\scriptsize \textbf{\coverage{30}{100}}} \\
{\scriptsize \hspace*{1.25em}hard (30)} & {\scriptsize \textbf{\coverage{30}{100}}} & {\scriptsize \textbf{\coverage{30}{100}}} & {\scriptsize \textbf{\coverage{30}{100}}} & {\scriptsize \coverage{24}{80}} & {\scriptsize \coverage{24}{80}} \\
\midrule
spanner (90) & \textbf{\coverage{90}{100}} & \textbf{\coverage{90}{100}} & \textbf{\coverage{90}{100}} & \coverage{73}{81} & \textbf{\coverage{90}{100}} \\
{\scriptsize \hspace*{1.25em}easy (30)} & {\scriptsize \textbf{\coverage{30}{100}}} & {\scriptsize \textbf{\coverage{30}{100}}} & {\scriptsize \textbf{\coverage{30}{100}}} & {\scriptsize \textbf{\coverage{30}{100}}} & {\scriptsize \textbf{\coverage{30}{100}}} \\
{\scriptsize \hspace*{1.25em}medium (30)} & {\scriptsize \textbf{\coverage{30}{100}}} & {\scriptsize \textbf{\coverage{30}{100}}} & {\scriptsize \textbf{\coverage{30}{100}}} & {\scriptsize \textbf{\coverage{30}{100}}} & {\scriptsize \textbf{\coverage{30}{100}}} \\
{\scriptsize \hspace*{1.25em}hard (30)} & {\scriptsize \textbf{\coverage{30}{100}}} & {\scriptsize \textbf{\coverage{30}{100}}} & {\scriptsize \textbf{\coverage{30}{100}}} & {\scriptsize \coverage{13}{43}} & {\scriptsize \textbf{\coverage{30}{100}}} \\
\midrule
satellite (90) & \textbf{\coverage{88}{98}} & \coverage{87}{97} & \coverage{63}{70} & \coverage{55}{61} & \coverage{51}{57} \\
{\scriptsize \hspace*{1.25em}easy (30)} & {\scriptsize \textbf{\coverage{30}{100}}} & {\scriptsize \textbf{\coverage{30}{100}}} & {\scriptsize \textbf{\coverage{30}{100}}} & {\scriptsize \textbf{\coverage{30}{100}}} & {\scriptsize \textbf{\coverage{30}{100}}} \\
{\scriptsize \hspace*{1.25em}medium (30)} & {\scriptsize \textbf{\coverage{30}{100}}} & {\scriptsize \textbf{\coverage{30}{100}}} & {\scriptsize \textbf{\coverage{30}{100}}} & {\scriptsize \coverage{22}{73}} & {\scriptsize \coverage{21}{70}} \\
{\scriptsize \hspace*{1.25em}hard (30)} & {\scriptsize \textbf{\coverage{28}{93}}} & {\scriptsize \coverage{27}{90}} & {\scriptsize \coverage{3}{10}} & {\scriptsize \coverage{3}{10}} & {\scriptsize \coverage{0}{0}} \\
\midrule
transport (90) & \coverage{86}{96} & \textbf{\coverage{87}{97}} & \coverage{73}{81} & \coverage{74}{82} & \coverage{74}{82} \\
{\scriptsize \hspace*{1.25em}easy (30)} & {\scriptsize \textbf{\coverage{30}{100}}} & {\scriptsize \textbf{\coverage{30}{100}}} & {\scriptsize \textbf{\coverage{30}{100}}} & {\scriptsize \textbf{\coverage{30}{100}}} & {\scriptsize \coverage{29}{97}} \\
{\scriptsize \hspace*{1.25em}medium (30)} & {\scriptsize \coverage{29}{97}} & {\scriptsize \textbf{\coverage{30}{100}}} & {\scriptsize \textbf{\coverage{30}{100}}} & {\scriptsize \coverage{25}{83}} & {\scriptsize \coverage{25}{83}} \\
{\scriptsize \hspace*{1.25em}hard (30)} & {\scriptsize \textbf{\coverage{27}{90}}} & {\scriptsize \textbf{\coverage{27}{90}}} & {\scriptsize \coverage{13}{43}} & {\scriptsize \coverage{19}{63}} & {\scriptsize \coverage{20}{67}} \\
\midrule
ferry (90) & \textbf{\coverage{80}{89}} & \textbf{\coverage{80}{89}} & \textbf{\coverage{80}{89}} & \coverage{64}{71} & \coverage{62}{69} \\
{\scriptsize \hspace*{1.25em}easy (30)} & {\scriptsize \textbf{\coverage{30}{100}}} & {\scriptsize \textbf{\coverage{30}{100}}} & {\scriptsize \textbf{\coverage{30}{100}}} & {\scriptsize \textbf{\coverage{30}{100}}} & {\scriptsize \coverage{29}{97}} \\
{\scriptsize \hspace*{1.25em}medium (30)} & {\scriptsize \textbf{\coverage{30}{100}}} & {\scriptsize \textbf{\coverage{30}{100}}} & {\scriptsize \textbf{\coverage{30}{100}}} & {\scriptsize \textbf{\coverage{30}{100}}} & {\scriptsize \coverage{29}{97}} \\
{\scriptsize \hspace*{1.25em}hard (30)} & {\scriptsize \textbf{\coverage{20}{67}}} & {\scriptsize \textbf{\coverage{20}{67}}} & {\scriptsize \textbf{\coverage{20}{67}}} & {\scriptsize \coverage{4}{13}} & {\scriptsize \coverage{4}{13}} \\
\midrule
rovers (90) & \textbf{\coverage{67}{74}} & \coverage{59}{66} & \coverage{51}{57} & \coverage{43}{48} & \coverage{27}{30} \\
{\scriptsize \hspace*{1.25em}easy (30)} & {\scriptsize \textbf{\coverage{30}{100}}} & {\scriptsize \textbf{\coverage{30}{100}}} & {\scriptsize \textbf{\coverage{30}{100}}} & {\scriptsize \textbf{\coverage{30}{100}}} & {\scriptsize \coverage{27}{90}} \\
{\scriptsize \hspace*{1.25em}medium (30)} & {\scriptsize \textbf{\coverage{24}{80}}} & {\scriptsize \coverage{17}{57}} & {\scriptsize \coverage{19}{63}} & {\scriptsize \coverage{13}{43}} & {\scriptsize \coverage{0}{0}} \\
{\scriptsize \hspace*{1.25em}hard (30)} & {\scriptsize \textbf{\coverage{13}{43}}} & {\scriptsize \coverage{12}{40}} & {\scriptsize \coverage{2}{7}} & {\scriptsize \coverage{0}{0}} & {\scriptsize \coverage{0}{0}} \\
\midrule
childsnack (90) & \coverage{64}{71} & \coverage{66}{73} & \textbf{\coverage{70}{78}} & \coverage{59}{66} & \coverage{41}{46} \\
{\scriptsize \hspace*{1.25em}easy (30)} & {\scriptsize \textbf{\coverage{30}{100}}} & {\scriptsize \textbf{\coverage{30}{100}}} & {\scriptsize \textbf{\coverage{30}{100}}} & {\scriptsize \textbf{\coverage{30}{100}}} & {\scriptsize \coverage{27}{90}} \\
{\scriptsize \hspace*{1.25em}medium (30)} & {\scriptsize \textbf{\coverage{30}{100}}} & {\scriptsize \textbf{\coverage{30}{100}}} & {\scriptsize \textbf{\coverage{30}{100}}} & {\scriptsize \coverage{29}{97}} & {\scriptsize \coverage{12}{40}} \\
{\scriptsize \hspace*{1.25em}hard (30)} & {\scriptsize \coverage{4}{13}} & {\scriptsize \coverage{6}{20}} & {\scriptsize \textbf{\coverage{10}{33}}} & {\scriptsize \coverage{0}{0}} & {\scriptsize \coverage{2}{7}} \\
\midrule
sokoban (90) & \coverage{13}{14} & \coverage{11}{12} & \coverage{12}{13} & \textbf{\coverage{14}{16}} & \textbf{\coverage{14}{16}} \\
{\scriptsize \hspace*{1.25em}easy (30)} & {\scriptsize \coverage{13}{43}} & {\scriptsize \coverage{11}{37}} & {\scriptsize \coverage{12}{40}} & {\scriptsize \textbf{\coverage{14}{47}}} & {\scriptsize \textbf{\coverage{14}{47}}} \\
{\scriptsize \hspace*{1.25em}medium (30)} & {\scriptsize \coverage{0}{0}} & {\scriptsize \coverage{0}{0}} & {\scriptsize \coverage{0}{0}} & {\scriptsize \coverage{0}{0}} & {\scriptsize \coverage{0}{0}} \\
{\scriptsize \hspace*{1.25em}hard (30)} & {\scriptsize \coverage{0}{0}} & {\scriptsize \coverage{0}{0}} & {\scriptsize \coverage{0}{0}} & {\scriptsize \coverage{0}{0}} & {\scriptsize \coverage{0}{0}} \\
\midrule
floortile (90) & \coverage{0}{0} & \coverage{0}{0} & \coverage{0}{0} & \coverage{0}{0} & \coverage{0}{0} \\
{\scriptsize \hspace*{1.25em}easy (30)} & {\scriptsize \coverage{0}{0}} & {\scriptsize \coverage{0}{0}} & {\scriptsize \coverage{0}{0}} & {\scriptsize \coverage{0}{0}} & {\scriptsize \coverage{0}{0}} \\
{\scriptsize \hspace*{1.25em}medium (30)} & {\scriptsize \coverage{0}{0}} & {\scriptsize \coverage{0}{0}} & {\scriptsize \coverage{0}{0}} & {\scriptsize \coverage{0}{0}} & {\scriptsize \coverage{0}{0}} \\
{\scriptsize \hspace*{1.25em}hard (30)} & {\scriptsize \coverage{0}{0}} & {\scriptsize \coverage{0}{0}} & {\scriptsize \coverage{0}{0}} & {\scriptsize \coverage{0}{0}} & {\scriptsize \coverage{0}{0}} \\
\midrule
\noalign{\vskip-\belowrulesep}
\rowcolor{black!8!white}
\btcoloredrowstrut\emph{Total (900)}
& \textbf{\coverage{668}{74}} & \coverage{659}{73} & \coverage{619}{69} & \coverage{542}{60} & \coverage{519}{58}  \\
\noalign{\vskip-\aboverulesep}
\bottomrule
\end{tabular}

    \caption{%
        IPC 2023 benchmark coverage results for AIW with AD encoding compared to BAIW, C-AIW, and Flat AD policies, as well as the baseline Flat AA (repeated from table \ref{tab:coverage:all}. Numbers in parentheses are percentages of row totals.}%
    \label{tab:coverage:riw-variants}
\end{table*}

\begin{table*}[t]
    \centering
    \footnotesize
    \renewcommand{\arraystretch}{0.73}
    \begin{tabular}{lrrrr}
\toprule
Domain & AIW & BAIW & C-AIW & Flat AD \\
\midrule
depots (86) & \textbf{\coverage{86}{100}} & \textbf{\coverage{86}{100}} & \textbf{\coverage{86}{100}} & \coverage{8}{9} \\
\midrule
delivery (80) & \textbf{\coverage{80}{100}} & \textbf{\coverage{80}{100}} & \textbf{\coverage{80}{100}} & \coverage{63}{79} \\
\midrule
driverlog (123) & \coverage{120}{98} & \coverage{122}{99} & \textbf{\coverage{123}{100}} & \coverage{120}{98} \\
\midrule
goldminer (359) & \textbf{\coverage{359}{100}} & \textbf{\coverage{359}{100}} & \textbf{\coverage{359}{100}} & \coverage{155}{43} \\
\midrule
grippers (312) & \textbf{\coverage{312}{100}} & \textbf{\coverage{312}{100}} & \textbf{\coverage{312}{100}} & \textbf{\coverage{312}{100}} \\
\midrule
freecell (117) & \textbf{\coverage{117}{100}} & \textbf{\coverage{117}{100}} & \textbf{\coverage{117}{100}} & \textbf{\coverage{117}{100}} \\
\midrule
visitall (100) & \textbf{\coverage{100}{100}} & \textbf{\coverage{100}{100}} & \textbf{\coverage{100}{100}} & \coverage{92}{92} \\
\midrule
reward (99) & \textbf{\coverage{99}{100}} & \textbf{\coverage{99}{100}} & \textbf{\coverage{99}{100}} & \coverage{95}{96} \\
\midrule
grid* (100) & \textbf{\coverage{100}{100}} & \textbf{\coverage{100}{100}} & \textbf{\coverage{100}{100}} & \coverage{24}{24} \\
\midrule
logistics* (100) & \textbf{\coverage{97}{97}} & \coverage{48}{48} & \coverage{93}{93} & \coverage{70}{70} \\
\midrule
satellite* (100) & \textbf{\coverage{100}{100}} & \textbf{\coverage{100}{100}} & \textbf{\coverage{100}{100}} & \coverage{54}{54} \\
\midrule
\noalign{\vskip-\belowrulesep}
\rowcolor{black!8!white}
\btcoloredrowstrut\emph{Total (1576) }
& \textbf{\coverage{1570}{100}} & \coverage{1523}{97} & \coverage{1569}{100} & \coverage{1110}{70}   \\
\noalign{\vskip-\aboverulesep}
\bottomrule
\end{tabular}
    \caption{%
        Benchmark coverage results for AIW compared to its variants BAIW, C-AIW, and Flat policy, all using AD encodings. Domains marked * are known to require logic beyond $C_2$ expressiveness. Numbers in parentheses are percentages of row totals.}%
    \label{tab:coverage:custom-ablation}
\end{table*}
\section{Model and Encoding Ablations}
We ablate the main model, \ie, AIW lookaheads coupled with AD encoding, along several structural changes in either encoding, IW lookahead behaviour, or both:
\begin{itemize}
    \item \emph{No Depthnodes} -- omitting depth nodes $o_d$ and associated depth atoms $\mathcal{R}_D^s$ from the encoding, resulting in the relation set $\mathcal{R}^\mathcal{T} \setminus \mathcal{R}^\mathcal{T}_\mathrm{D}$.
    \item \emph{No Tree-Atoms} -- omitting the encoding step for the parent relation atoms, resulting in the relation set $\mathcal{R}^\mathcal{T} \setminus \mathcal{R}^\mathcal{T}_\mathrm{E}$.
    \item \emph{No Depth/Tree-Atoms} -- omitting both depth and tree atoms, resulting in the relation set $\mathcal{R}^\mathcal{T} \setminus (\mathcal{R}_D^\mathcal{T} \cup \mathcal{R}^\mathcal{T}_\mathrm{E})$.
    \item \emph{Reduced Goals} -- Full AD relation set $\mathcal{R}^\mathcal{T}$ encoded, yet AIW now also applies the abstracted novelty checks on atoms $P(o_1, \dots, o_n)$ that match an atom $P_g(o_1, \dots, o_n)$ in the goal.
\end{itemize}
Given that Tables~\ref{tab:coverage:all} and~\ref{tab:coverage:riw-variants} already cover several ablation directions, Table~\ref{tab:coverage:main-ablation} isolates the remaining AD-encoding components. Most of these components were introduced to stabilize test-time behavior. During development, we often observed large variance in test outcomes, which decreased after the full encoding was introduced. The ablation results show no large aggregate performance drop for most removals. However, they indicate policy instabilities: some easy test instances remain unsolved, even when most hard instances are solved, as in Spanner when depth nodes are removed. Values also fluctuate between the ablation setups, resulting in better outcomes, although more components have been deactivated. The only consistently high-impact ablation is the abstraction of atoms that also occur in the goal. Removing this exception substantially reduces coverage on the harder instances of most domains.

\begin{table*}[t]
    \centering
    \footnotesize
    \setlength{\tabcolsep}{6pt}
    \renewcommand{\arraystretch}{0.73}
    \begin{tabular}{lrrrrr}
\toprule
Domain & AIW & No Depthnodes & No Tree-Atoms & No Depthnodes & Reduced Goals \\
&&&& No Tree-Atoms & \\
\midrule
blocksworld (90) & \coverage{90}{100} & \textbf{\coverage{90}{100}} & \coverage{84}{93} & \coverage{84}{93} & \coverage{78}{87} \\
\hspace*{1.25em}easy (30) & \textbf{\coverage{30}{100}} & \textbf{\coverage{30}{100}} & \textbf{\coverage{30}{100}} & \coverage{28}{93} & \textbf{\coverage{30}{100}} \\
\hspace*{1.25em}medium (30) & \textbf{\coverage{30}{100}} & \textbf{\coverage{30}{100}} & \textbf{\coverage{30}{100}} & \coverage{29}{97} & \textbf{\coverage{30}{100}} \\
\hspace*{1.25em}hard (30) & \coverage{30}{100} & \textbf{\coverage{30}{100}} & \coverage{24}{80} & \coverage{27}{90} & \coverage{18}{60} \\
\midrule
miconic (90) & \textbf{\coverage{90}{100}} & \textbf{\coverage{90}{100}} & \textbf{\coverage{90}{100}} & \textbf{\coverage{90}{100}} & \textbf{\coverage{90}{100}} \\
\hspace*{1.25em}easy (30) & \textbf{\coverage{30}{100}} & \textbf{\coverage{30}{100}} & \textbf{\coverage{30}{100}} & \textbf{\coverage{30}{100}} & \textbf{\coverage{30}{100}} \\
\hspace*{1.25em}medium (30) & \textbf{\coverage{30}{100}} & \textbf{\coverage{30}{100}} & \textbf{\coverage{30}{100}} & \textbf{\coverage{30}{100}} & \textbf{\coverage{30}{100}} \\
\hspace*{1.25em}hard (30) & \textbf{\coverage{30}{100}} & \textbf{\coverage{30}{100}} & \textbf{\coverage{30}{100}} & \textbf{\coverage{30}{100}} & \textbf{\coverage{30}{100}} \\
\midrule
spanner (90) & \textbf{\coverage{90}{100}} & \coverage{86}{96} & \coverage{81}{90} & \coverage{87}{97} & \textbf{\coverage{90}{100}} \\
\hspace*{1.25em}easy (30) & \textbf{\coverage{30}{100}} & \coverage{27}{90} & \textbf{\coverage{30}{100}} & \textbf{\coverage{30}{100}} & \textbf{\coverage{30}{100}} \\
\hspace*{1.25em}medium (30) & \textbf{\coverage{30}{100}} & \textbf{\coverage{30}{100}} & \textbf{\coverage{30}{100}} & \textbf{\coverage{30}{100}} & \textbf{\coverage{30}{100}} \\
\hspace*{1.25em}hard (30) & \textbf{\coverage{30}{100}} & \coverage{29}{97} & \coverage{21}{70} & \coverage{27}{90} & \textbf{\coverage{30}{100}} \\
\midrule
satellite (90) & \textbf{\coverage{88}{98}} & \textbf{\coverage{88}{98}} & \coverage{84}{93} & \coverage{85}{94} & \coverage{64}{71} \\
\hspace*{1.25em}easy (30) & \textbf{\coverage{30}{100}} & \textbf{\coverage{30}{100}} & \textbf{\coverage{30}{100}} & \textbf{\coverage{30}{100}} & \textbf{\coverage{30}{100}} \\
\hspace*{1.25em}medium (30) & \textbf{\coverage{30}{100}} & \textbf{\coverage{30}{100}} & \textbf{\coverage{30}{100}} & \coverage{29}{97} & \coverage{28}{93} \\
\hspace*{1.25em}hard (30) & \textbf{\coverage{28}{93}} & \textbf{\coverage{28}{93}} & \coverage{24}{80} & \coverage{26}{87} & \coverage{6}{20} \\
\midrule
transport (90) & \textbf{\coverage{86}{96}} & \coverage{83}{92} & \coverage{82}{91} & \textbf{\coverage{86}{96}} & \coverage{63}{70} \\
\hspace*{1.25em}easy (30) & \textbf{\coverage{30}{100}} & \textbf{\coverage{30}{100}} & \textbf{\coverage{30}{100}} & \textbf{\coverage{30}{100}} & \textbf{\coverage{30}{100}} \\
\hspace*{1.25em}medium (30) & \textbf{\coverage{29}{97}} & \coverage{28}{93} & \coverage{27}{90} & \textbf{\coverage{29}{97}} & \coverage{26}{87} \\
\hspace*{1.25em}hard (30) & \textbf{\coverage{27}{90}} & \coverage{25}{83} & \coverage{25}{83} & \textbf{\coverage{27}{90}} & \coverage{7}{23} \\
\midrule
ferry (90) & \textbf{\coverage{80}{89}} & \textbf{\coverage{80}{89}} & \textbf{\coverage{80}{89}} & \textbf{\coverage{80}{89}} & \coverage{69}{77} \\
\hspace*{1.25em}easy (30) & \textbf{\coverage{30}{100}} & \textbf{\coverage{30}{100}} & \textbf{\coverage{30}{100}} & \textbf{\coverage{30}{100}} & \textbf{\coverage{30}{100}} \\
\hspace*{1.25em}medium (30) & \textbf{\coverage{30}{100}} & \textbf{\coverage{30}{100}} & \textbf{\coverage{30}{100}} & \textbf{\coverage{30}{100}} & \textbf{\coverage{30}{100}} \\
\hspace*{1.25em}hard (30) & \textbf{\coverage{20}{67}} & \textbf{\coverage{20}{67}} & \textbf{\coverage{20}{67}} & \textbf{\coverage{20}{67}} & \coverage{9}{30} \\
\midrule
rovers (90) & \textbf{\coverage{67}{74}} & \coverage{57}{63} & \coverage{43}{48} & \coverage{56}{62} & \coverage{36}{40} \\
\hspace*{1.25em}easy (30) & \textbf{\coverage{30}{100}} & \textbf{\coverage{30}{100}} & \textbf{\coverage{30}{100}} & \textbf{\coverage{30}{100}} & \coverage{29}{97} \\
\hspace*{1.25em}medium (30) & \textbf{\coverage{24}{80}} & \coverage{20}{67} & \coverage{9}{30} & \coverage{19}{63} & \coverage{7}{23} \\
\hspace*{1.25em}hard (30) & \textbf{\coverage{13}{43}} & \coverage{7}{23} & \coverage{4}{13} & \coverage{7}{23} & \coverage{0}{0} \\
\midrule
childsnack (90) & \coverage{64}{71} & \textbf{\coverage{66}{73}} & \coverage{62}{69} & \textbf{\coverage{66}{73}} & \coverage{65}{72} \\
\hspace*{1.25em}easy (30) & \textbf{\coverage{30}{100}} & \textbf{\coverage{30}{100}} & \textbf{\coverage{30}{100}} & \textbf{\coverage{30}{100}} & \textbf{\coverage{30}{100}} \\
\hspace*{1.25em}medium (30) & \textbf{\coverage{30}{100}} & \textbf{\coverage{30}{100}} & \textbf{\coverage{30}{100}} & \textbf{\coverage{30}{100}} & \textbf{\coverage{30}{100}} \\
\hspace*{1.25em}hard (30) & \coverage{4}{13} & \textbf{\coverage{6}{20}} & \coverage{2}{7} & \textbf{\coverage{6}{20}} & \coverage{5}{17} \\
\midrule
sokoban (90) & \coverage{13}{14} & \coverage{13}{14} & \textbf{\coverage{14}{16}} & \coverage{11}{12} & \coverage{11}{12} \\
\hspace*{1.25em}easy (30) & \coverage{13}{43} & \coverage{13}{43} & \textbf{\coverage{14}{47}} & \coverage{11}{37} & \coverage{11}{37} \\
\hspace*{1.25em}medium (30) & \coverage{0}{0} & \coverage{0}{0} & \coverage{0}{0} & \coverage{0}{0} & \coverage{0}{0} \\
\hspace*{1.25em}hard (30) & \coverage{0}{0} & \coverage{0}{0} & \coverage{0}{0} & \coverage{0}{0} & \coverage{0}{0} \\
\midrule
floortile (90) & \coverage{0}{0} & \coverage{0}{0} & \coverage{0}{0} & \coverage{0}{0} & \coverage{0}{0} \\
\hspace*{1.25em}easy (30) & \coverage{0}{0} & \coverage{0}{0} & \coverage{0}{0} & \coverage{0}{0} & \coverage{0}{0} \\
\hspace*{1.25em}medium (30) & \coverage{0}{0} & \coverage{0}{0} & \coverage{0}{0} & \coverage{0}{0} & \coverage{0}{0} \\
\hspace*{1.25em}hard (30) & \coverage{0}{0} & \coverage{0}{0} & \coverage{0}{0} & \coverage{0}{0} & \coverage{0}{0} \\
\midrule
\noalign{\vskip-\belowrulesep}
\rowcolor{black!8!white}
\btcoloredrowstrut\emph{Total (900)}
& \textbf{\coverage{668}{74}} & \coverage{653}{73} & \coverage{620}{69} & \coverage{645}{72} & \coverage{566}{63} \\
\noalign{\vskip-\aboverulesep}
\bottomrule
\end{tabular}

    \caption{%
        IPC 2023 benchmark ablation coverage results for AIW with AD encoding. Numbers in parentheses are percentages of row totals.}%
    \label{tab:coverage:main-ablation}
\end{table*}

\paragraph{Encoding Ablations.}
As mentioned in Section~\ref{sec:encodings}, we consider two interpolating encodings: Internal and Internal-Delta. External encodings represent each successor with its full relation set in a separate graph and process state transitions sequentially. At the other end, Aggregated-Delta (AD) represents successors only through their delta atoms and encodes all transitions jointly. Internal and Internal-Delta therefore isolate whether performance gains arise from merging current and successor states into a single graph, from representing successors through state differences, or from combining both mechanisms.

\begin{table*}[th]
    \centering
    \footnotesize
    \setlength{\tabcolsep}{6pt}
    \renewcommand{\arraystretch}{0.73}
    \begin{tabular}{lrrrr}
\toprule
Domain & AD & ID & Int & Ext \\
\midrule
blocksworld (90) & \textbf{\coverage{90}{100}} & \coverage{75}{83} & \coverage{76}{84} & \coverage{78}{87} \\
\hspace*{1.25em}easy (30) & \textbf{\coverage{30}{100}} & \textbf{\coverage{30}{100}} & \textbf{\coverage{30}{100}} & \textbf{\coverage{30}{100}} \\
\hspace*{1.25em}medium (30) & \textbf{\coverage{30}{100}} & \textbf{\coverage{30}{100}} & \textbf{\coverage{30}{100}} & \textbf{\coverage{30}{100}} \\
\hspace*{1.25em}hard (30) & \textbf{\coverage{30}{100}} & \coverage{15}{50} & \coverage{16}{53} & \coverage{18}{60} \\
\midrule
miconic (90) & \textbf{\coverage{90}{100}} & \coverage{72}{80} & \coverage{69}{77} & \coverage{72}{80} \\
\hspace*{1.25em}easy (30) & \textbf{\coverage{30}{100}} & \textbf{\coverage{30}{100}} & \textbf{\coverage{30}{100}} & \textbf{\coverage{30}{100}} \\
\hspace*{1.25em}medium (30) & \textbf{\coverage{30}{100}} & \textbf{\coverage{30}{100}} & \textbf{\coverage{30}{100}} & \textbf{\coverage{30}{100}} \\
\hspace*{1.25em}hard (30) & \textbf{\coverage{30}{100}} & \coverage{12}{40} & \coverage{9}{30} & \coverage{12}{40} \\
\midrule
spanner (90) & \textbf{\coverage{90}{100}} & \coverage{30}{33} & \coverage{32}{36} & \coverage{32}{36} \\
\hspace*{1.25em}easy (30) & \textbf{\coverage{30}{100}} & \textbf{\coverage{30}{100}} & \textbf{\coverage{30}{100}} & \textbf{\coverage{30}{100}} \\
\hspace*{1.25em}medium (30) & \textbf{\coverage{30}{100}} & \coverage{0}{0} & \coverage{2}{7} & \coverage{2}{7} \\
\hspace*{1.25em}hard (30) & \textbf{\coverage{30}{100}} & \coverage{0}{0} & \coverage{0}{0} & \coverage{0}{0} \\
\midrule
satellite (90) & \textbf{\coverage{88}{98}} & \coverage{64}{71} & \coverage{48}{53} & \coverage{61}{68} \\
\hspace*{1.25em}easy (30) & \textbf{\coverage{30}{100}} & \textbf{\coverage{30}{100}} & \textbf{\coverage{30}{100}} & \textbf{\coverage{30}{100}} \\
\hspace*{1.25em}medium (30) & \textbf{\coverage{30}{100}} & \textbf{\coverage{30}{100}} & \coverage{16}{53} & \textbf{\coverage{30}{100}} \\
\hspace*{1.25em}hard (30) & \textbf{\coverage{28}{93}} & \coverage{4}{13} & \coverage{2}{7} & \coverage{1}{3} \\
\midrule
transport (90) & \textbf{\coverage{86}{96}} & \coverage{64}{71} & \coverage{45}{50} & \coverage{55}{61} \\
\hspace*{1.25em}easy (30) & \textbf{\coverage{30}{100}} & \textbf{\coverage{30}{100}} & \textbf{\coverage{30}{100}} & \coverage{29}{97} \\
\hspace*{1.25em}medium (30) & \textbf{\coverage{29}{97}} & \coverage{28}{93} & \coverage{14}{47} & \coverage{19}{63} \\
\hspace*{1.25em}hard (30) & \textbf{\coverage{27}{90}} & \coverage{6}{20} & \coverage{1}{3} & \coverage{7}{23} \\
\midrule
ferry (90) & \textbf{\coverage{80}{89}} & \coverage{69}{77} & \coverage{69}{77} & \coverage{71}{79} \\
\hspace*{1.25em}easy (30) & \textbf{\coverage{30}{100}} & \textbf{\coverage{30}{100}} & \textbf{\coverage{30}{100}} & \textbf{\coverage{30}{100}} \\
\hspace*{1.25em}medium (30) & \textbf{\coverage{30}{100}} & \textbf{\coverage{30}{100}} & \textbf{\coverage{30}{100}} & \textbf{\coverage{30}{100}} \\
\hspace*{1.25em}hard (30) & \textbf{\coverage{20}{67}} & \coverage{9}{30} & \coverage{9}{30} & \coverage{11}{37} \\
\midrule
rovers (90) & \textbf{\coverage{67}{74}} & \coverage{29}{32} & \coverage{27}{30} & \coverage{24}{27} \\
\hspace*{1.25em}easy (30) & \textbf{\coverage{30}{100}} & \coverage{29}{97} & \coverage{27}{90} & \coverage{24}{80} \\
\hspace*{1.25em}medium (30) & \textbf{\coverage{24}{80}} & \coverage{0}{0} & \coverage{0}{0} & \coverage{0}{0} \\
\hspace*{1.25em}hard (30) & \textbf{\coverage{13}{43}} & \coverage{0}{0} & \coverage{0}{0} & \coverage{0}{0} \\
\midrule
childsnack (90) & \textbf{\coverage{64}{71}} & \coverage{32}{36} & \coverage{38}{42} & \coverage{47}{52} \\
\hspace*{1.25em}easy (30) & \textbf{\coverage{30}{100}} & \coverage{20}{67} & \textbf{\coverage{30}{100}} & \textbf{\coverage{30}{100}} \\
\hspace*{1.25em}medium (30) & \textbf{\coverage{30}{100}} & \coverage{12}{40} & \coverage{8}{27} & \coverage{17}{57} \\
\hspace*{1.25em}hard (30) & \textbf{\coverage{4}{13}} & \coverage{0}{0} & \coverage{0}{0} & \coverage{0}{0} \\
\midrule
sokoban (90) & \coverage{13}{14} & \coverage{12}{13} & \coverage{7}{8} & \textbf{\coverage{17}{19}} \\
\hspace*{1.25em}easy (30) & \coverage{13}{43} & \coverage{12}{40} & \coverage{7}{23} & \textbf{\coverage{17}{57}} \\
\hspace*{1.25em}medium (30) & \coverage{0}{0} & \coverage{0}{0} & \coverage{0}{0} & \coverage{0}{0} \\
\hspace*{1.25em}hard (30) & \coverage{0}{0} & \coverage{0}{0} & \coverage{0}{0} & \coverage{0}{0} \\
\midrule
floortile (90) & \coverage{0}{0} & \coverage{0}{0} & \coverage{0}{0} & \coverage{0}{0} \\
\hspace*{1.25em}easy (30) & \coverage{0}{0} & \coverage{0}{0} & \coverage{0}{0} & \coverage{0}{0} \\
\hspace*{1.25em}medium (30) & \coverage{0}{0} & \coverage{0}{0} & \coverage{0}{0} & \coverage{0}{0} \\
\hspace*{1.25em}hard (30) & \coverage{0}{0} & \coverage{0}{0} & \coverage{0}{0} & \coverage{0}{0} \\
\midrule
\noalign{\vskip-\belowrulesep}
\rowcolor{black!8!white}
\btcoloredrowstrut\emph{Total (900)}
& \textbf{\coverage{668}{74}} & \coverage{447}{50} & \coverage{411}{46} & \coverage{457}{51} \\
\noalign{\vskip-\aboverulesep}
\bottomrule
\end{tabular}

    \caption{%
        IPC 2023 benchmark ablation coverage results for encodings External, Internal, Internal-Delta, and Aggregated-Delta paired with AIW lookaheads. Numbers in parentheses are percentages of row totals.}%
    \label{tab:coverage:main-encoding-ablation}
\end{table*}

\begin{table*}[th]
    \centering
    \footnotesize
    \setlength{\tabcolsep}{6pt}
    \renewcommand{\arraystretch}{0.73}
    \begin{tabular}{lrrrr}
\toprule
Domain & AD & ID & Int & Ext \\
\midrule
depots (86) & \textbf{\coverage{86}{100}} & \coverage{79}{92} & \coverage{83}{97} & \coverage{83}{97} \\
\midrule
delivery (80) & \textbf{\coverage{80}{100}} & \textbf{\coverage{80}{100}} & \coverage{75}{94} & \textbf{\coverage{80}{100}} \\
\midrule
driverlog (123) & \coverage{120}{98} & \textbf{\coverage{122}{99}} & \coverage{116}{94} & \coverage{116}{94} \\
\midrule
goldminer (359) & \textbf{\coverage{359}{100}} & \coverage{317}{88} & \coverage{50}{14} & \coverage{133}{37} \\
\midrule
grippers (312) & \textbf{\coverage{312}{100}} & \textbf{\coverage{312}{100}} & \textbf{\coverage{312}{100}} & \textbf{\coverage{312}{100}} \\
\midrule
freecell (117) & \textbf{\coverage{117}{100}} & \textbf{\coverage{117}{100}} & \textbf{\coverage{117}{100}} & \textbf{\coverage{117}{100}} \\
\midrule
visitall (100) & \textbf{\coverage{100}{100}} & \coverage{98}{98} & \coverage{94}{94} & \textbf{\coverage{100}{100}} \\
\midrule
reward (99) & \textbf{\coverage{99}{100}} & \textbf{\coverage{99}{100}} & \coverage{86}{87} & \textbf{\coverage{99}{100}} \\
\midrule
grid* (100) & \textbf{\coverage{100}{100}} & \coverage{91}{91} & \coverage{90}{90} & \textbf{\coverage{100}{100}} \\
\midrule
logistics* (100) & \textbf{\coverage{97}{97}} & \coverage{69}{69} & \coverage{7}{7} & \coverage{0}{0} \\
\midrule
satellite* (100) & \textbf{\coverage{100}{100}} & \coverage{87}{87} & \coverage{19}{19} & \coverage{26}{26} \\
\midrule
\noalign{\vskip-\belowrulesep}
\rowcolor{black!8!white}
\btcoloredrowstrut\emph{Total (1576) }
& \textbf{\coverage{1570}{100}} & \coverage{1471}{93} & \coverage{1049}{67} & \coverage{1166}{74} \\
\noalign{\vskip-\aboverulesep}
\bottomrule
\end{tabular}

    \caption{%
        Encoding ablation coverage results for encodings External, Internal, Internal-Delta, and Aggregated-Delta paired with AIW lookaheads on our extended benchmark set. Numbers in parentheses are percentages of row totals.}%
    \label{tab:coverage:main-encoding-ablation-custom}
\end{table*}

Tables~\ref{tab:coverage:main-encoding-ablation} and~\ref{tab:coverage:main-encoding-ablation-custom} show that merging current and successor states into a single graph is not sufficient on its own. Internal encodings are consistently worse than External encodings, both in coverage and resource usage. This is likely due to the duplicated relation set, which effectively doubles the number of predicate-specific modules relative to External encodings and increases the number of messages aggregated by the R-GNN. Internal-Delta recovers much of the lost performance while substantially reducing memory usage, showing that explicit state differences are important for exploiting the structure exposed by IW lookaheads. This effect is particularly clear on the extended benchmark: External and Internal encodings fail on several domains where Internal-Delta achieves high coverage, including domains that require logic beyond $C_2$. In particular, Logistics is solved substantially more often, Satellite becomes almost fully solvable, and Goldminer is solved well only by Internal-Delta and AD. Nevertheless, Internal-Delta still remains below AD, and only AD achieves full coverage on the extended benchmark. These results indicate that both components are needed: compact delta-based successor representations and the joint encoding of the full transition set.

\section{IPC 2023 Benchmark Data Scaling Behavior}

\begin{figure}
    \centering
    \includegraphics[width=1\linewidth]{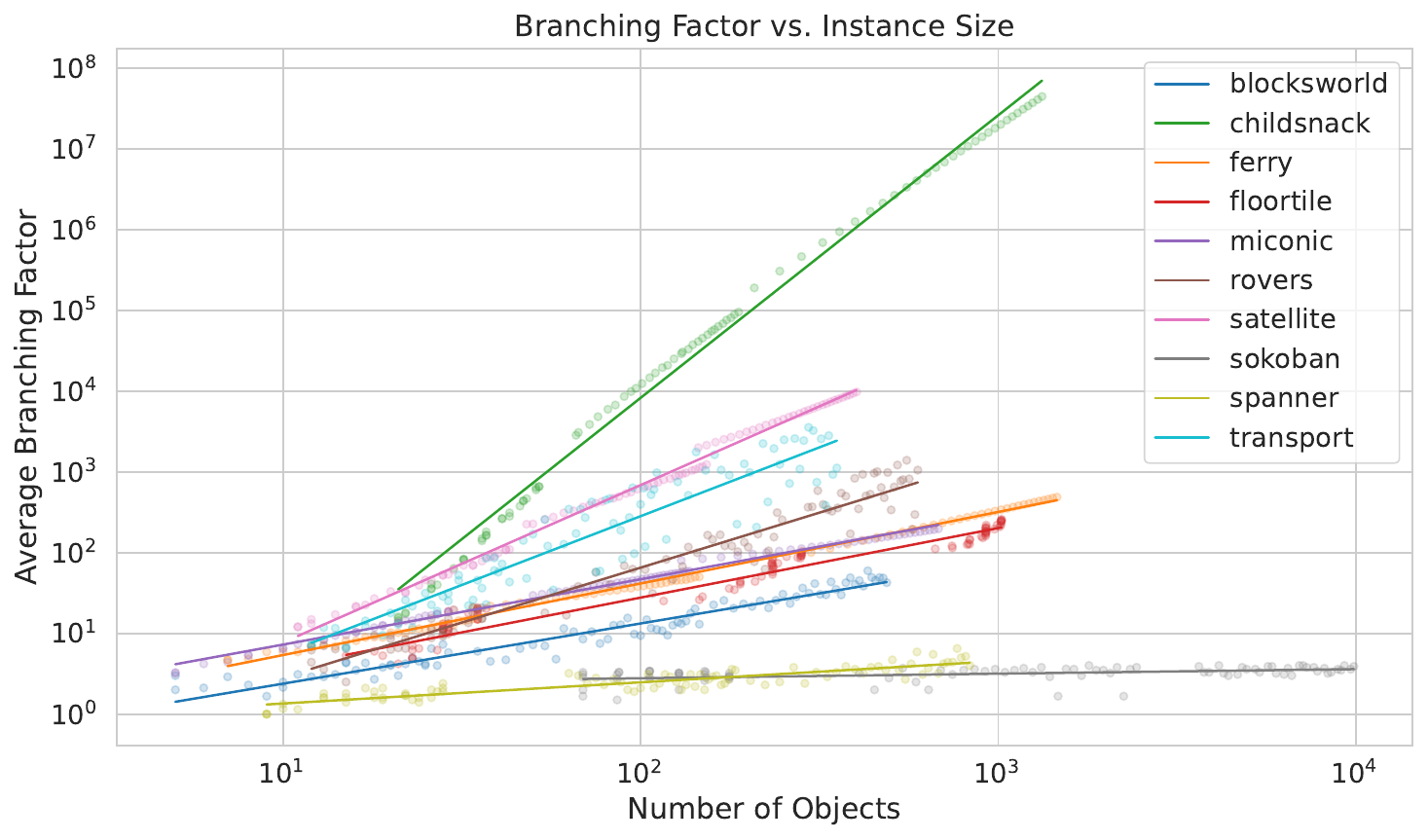}
\caption{Log-log plot of the branching factor for each IPC benchmark domain as a function of the number of objects.}
    \label{fig:branching-factor}
\end{figure}

One way to analyze problem difficulty is to examine how the branching factor scales with instance size. Since the size of the search tree grows exponentially in the branching factor, instances with larger branching factors are generally harder for search-based approaches. We estimate the average branching factor of an instance by running a random walk of length $10$ from the initial state, recording the number of applicable actions in each visited state, and averaging these counts. Figure~\ref{fig:branching-factor} shows the resulting scaling behavior for the IPC domains on a log-log scale. The plot highlights the particular difficulty of Childsnack, whose branching factor grows much faster than in the other domains and reaches values several orders of magnitude larger. Concretely, in Childsnack's hard category, instances \#1, \#10, \#20, and \#30 exhibit $0.25$, $3.7$, $17$, and $47$ million applicable actions in the initial state, respectively, which scale from requiring $1$ second up to $10$ minutes simply to generate. This, despite highly optimized software and latest hardware. 

Of course, branching factors alone do not explain all of a domain's difficulty. For example, the branching factor in Sokoban remains nearly constant with instance size, although the domain remains challenging for search due to complex long-term consequences of individual action choices.

\section{Domain Descriptions}
This appendix section complements the main text with supplementary content cataloguing every benchmark domain and summarizing the instance statistics highlighted in Table \ref{tab:data}. 
\begin{table*}[ht!]
    \centering
    \footnotesize
    \setlength{\tabcolsep}{4pt}
    \begin{tabular}{l rrrrr rrrrr rrrrr}
\toprule
 & \multicolumn{5}{c}{Train} & \multicolumn{5}{c}{Validation} & \multicolumn{5}{c}{Test} \\
\cmidrule(lr){2-6} \cmidrule(lr){7-11} \cmidrule(lr){12-16}
& & \multicolumn{2}{c}{Objects} & \multicolumn{2}{c}{Atoms} & & \multicolumn{2}{c}{Objects} & \multicolumn{2}{c}{Atoms} & & \multicolumn{2}{c}{Objects} & \multicolumn{2}{c}{Atoms} \\
\cmidrule(lr){3-4} \cmidrule(lr){5-6} \cmidrule(lr){8-9} \cmidrule(lr){10-11} \cmidrule(lr){13-14} \cmidrule(lr){15-16}
Domain & $\#$ & Min & Max & Min & Max & $\#$ & Min & Max & Min & Max & $\#$ & Min & Max & Min & Max \\
\midrule
delivery & 165 & 6 & 70 & 31 & 447 & 30 & 88 & 129 & 567 & 844 & 80 & 153 & 634 & 1006 & 4321 \\
depots & 589 & 13 & 20 & 62 & 104 & 50 & 21 & 25 & 100 & 130 & 86 & 26 & 60 & 126 & 313 \\
driverlog & 108 & 7 & 15 & 27 & 83 & 48 & 16 & 18 & 71 & 104 & 123 & 19 & 50 & 91 & 291 \\
freecell & 787 & 29 & 48 & 124 & 213 & 40 & 31 & 51 & 131 & 225 & 117 & 40 & 56 & 174 & 242 \\
goldminer & 190 & 4 & 20 & 25 & 121 & 40 & 4 & 238 & 25 & 1609 & 359 & 18 & 361 & 109 & 2456 \\
grippers & 113 & 7 & 42 & 17 & 123 & 40 & 8 & 126 & 19 & 309 & 312 & 22 & 127 & 47 & 308 \\
reward & 155 & 4 & 144 & 26 & 1098 & 45 & 169 & 225 & 1293 & 1732 & 99 & 256 & 841 & 1968 & 6543 \\
visitall & 99 & 4 & 144 & 22 & 962 & 27 & 169 & 225 & 1133 & 1517 & 100 & 256 & 900 & 1730 & 6182 \\
grid & 128 & 11 & 76 & 48 & 459 & 47 & 108 & 206 & 684 & 1349 & 100 & 278 & 890 & 1792 & 6027 \\
logistics & 96 & 6 & 26 & 18 & 78 & 40 & 30 & 52 & 90 & 156 & 100 & 60 & 120 & 180 & 360 \\
satellite & 59 & 6 & 34 & 18 & 158 & 66 & 53 & 122 & 216 & 644 & 100 & 140 & 435 & 804 & 3425 \\
\midrule
\midrule
blocksworld-ipc & 69 & 2 & 20 & 6 & 45 & 30 & 21 & 29 & 45 & 64 & 90 & 5 & 488 & 13 & 1019 \\
childsnack-ipc & 68 & 7 & 41 & 20 & 135 & 30 & 41 & 52 & 135 & 174 & 90 & 21 & 1327 & 63 & 4563 \\
ferry-ipc & 69 & 3 & 25 & 9 & 65 & 30 & 26 & 35 & 68 & 92 & 90 & 7 & 1461 & 18 & 3898 \\
floortile-ipc & 67 & 5 & 40 & 17 & 238 & 30 & 37 & 45 & 215 & 267 & 90 & 15 & 1022 & 79 & 6884 \\
miconic-ipc & 69 & 3 & 21 & 10 & 148 & 30 & 22 & 30 & 164 & 271 & 90 & 5 & 681 & 19 & 21443 \\
rovers-ipc & 69 & 10 & 28 & 35 & 254 & 30 & 28 & 36 & 193 & 422 & 90 & 12 & 596 & 56 & 294553 \\
satellite-ipc & 69 & 5 & 32 & 15 & 133 & 30 & 34 & 43 & 139 & 196 & 90 & 11 & 402 & 37 & 2402 \\
sokoban-ipc & 69 & 54 & 128 & 134 & 517 & 30 & 128 & 177 & 475 & 793 & 90 & 69 & 9884 & 219 & 55872 \\
spanner-ipc & 59 & 6 & 22 & 22 & 88 & 30 & 22 & 28 & 88 & 114 & 90 & 9 & 833 & 31 & 3961 \\
transport-ipc & 69 & 6 & 34 & 20 & 258 & 30 & 34 & 45 & 140 & 382 & 90 & 12 & 354 & 47 & 7785 \\
\bottomrule
\end{tabular}
    \caption{Statistics of the benchmarks used in our experiments. For each
    benchmark, we report the number of instances, as well as the minimum and
    maximum number of objects and atoms.
    }%
    \label{tab:data}
\end{table*}

\paragraph{Blocksworld}
In Blocksworld, the goal is to stack blocks on top of each other to form specific configurations. The actions involve picking up and putting down blocks, where blocks can only be picked up if they have no other blocks on top of them.

\paragraph{Childsnack}
In this domain, the task is to serve snacks to children. Some children are gluten allergic and need to be served gluten-free bread. One general policy is to always serve gluten-free bread to gluten allergic children first, then serve the rest to those that are left. Actions include making sandwiches and serving them to children.

\paragraph{Delivery}
In Delivery, the objective is to deliver packages to their respective destinations using a single truck. The truck can only carry one package at a time, and the actions include loading and unloading packages, as well as driving the truck between locations.

\paragraph{Depots}
The task in Depots is to move crates onto pallets. The instances contain multiple trucks and hoists that can be used to move crates. The actions include loading and unloading crates onto and from pallets, as well as driving trucks and operating hoists.

\paragraph{Driverlog}
In Driverlog, the goal is to transport packages and trucks to different locations. In comparison to Delivery and Logistics, this domain also includes drivers that are required to drive the trucks. The actions include loading and unloading packages onto and from trucks, as well as driving trucks and walking drivers to different locations.

\paragraph{Freecell}
The Freecell domain is based on a card game similar to Solitaire, where the objective is to move all cards to the locations specified in the goal. Actions are to move cards between different cells.

\paragraph{Ferry}
The Ferry domain is about transporting cars with a ferry. The ferry can only hold a single car at the time, and the goal is to move all cars to their respective destinations. Actions include loading and unloading cars onto and from the ferry, as well as sailing between shores.

\paragraph{Floortile}
In Floortile, the objective is to paint floor tiles in a specific pattern using a robot. The robot can move around the grid of tiles and paint them in different colors, but can not move to a painted tile. The challenge is to ensure that the final configuration of painted tiles matches the desired pattern, without getting stuck.

\paragraph{Goldminer}
The goal of Goldminer is for a robot to reach a location that contains gold. The environment is grid based and contains items such as bombs and laser cannons. Actions include moving and using the items to clear a path to the gold.

\paragraph{Grid}
The task in Grid is to navigate a grid and put keys in specific positions. Some positions are locked and require a key to open. The actions include moving in the grid, picking up keys, and unlocking locked positions.

\paragraph{Grippers}
In Gripper, the objective is to move balls from one room to another using a robot with two grippers. The robot can carry up to two balls at a time, and the actions include moving between rooms, picking up balls, and dropping them off in the target room.
Grippers is similar to Grippers, but instead of a single robot, there are multiple robots, and there can be more than two rooms.

\paragraph{Logistics}
In Logistics, the goal is to transport packages between different locations using trucks and airplanes. Trucks can only move within a city, while airplanes can fly between cities. The actions include loading and unloading packages onto and from vehicles, as well as driving trucks and flying airplanes to different locations.

\paragraph{Miconic}
Miconic is about transporting passengers in an elevator to their desired floors. The elevator can move up and down, and passengers can board and leave the elevator at different floors. The goal is to efficiently transport all passengers to their respective floors while minimizing the number of stops.

\paragraph{Reward}
The goal in Reward is to collect rewards that are placed in different cells. The environment is grid-based and contains obstacles that the agent can not pass. The actions include moving in the grid and collecting rewards.

\paragraph{Rovers}
Here rovers are tasked with exploring a planetary surface, collecting samples, and taking images. The rovers can move around the terrain, collect samples from specific locations, and take images of points of interest. Different rovers can traverse different types of terrain, which adds complexity to the planning task. The goal is to communicate data from different waypoints.

\paragraph{Satellite}
In Satellite, the goal is to capture images and point satellites in different directions. The actions include turning the satellites, taking images and calibrating instruments. The challenge is to manage the limited resources of the satellites while ensuring that all required images are captured.

\paragraph{Sokoban}
This domain is based on the classic puzzle game Sokoban, where the objective is to push boxes onto designated goal locations within a warehouse. The player can move in four directions and push boxes, but cannot pull them. The challenge lies in planning moves carefully to avoid getting boxes stuck in corners or against walls.

\paragraph{Spanner}
The task in Spanner is to tighten loose nuts using a wrench (spanner). The actions are to move, pickup spanners and tighten nuts. Whenever a wrench is used, it can not be used again. The difficulty is that the agent can only move in one direction, so if it missed to pick up a wrench that it later needs, it will not be able to solve the task.

\paragraph{Transport}
In Transport, the task is to drive packages to specific locations. Different vehicles have different capacities, so they can carry different number of packages. The actions include loading and unloading packages, as well as driving vehicles between locations.

\paragraph{Visitall}
Visitall is another grid based domain where the objective is to visit all the cells specified in the goal. The only action is to move between connected cells in the grid.

\subsection{Domain statistics}
Table \ref{tab:data} summarizes how each benchmark grows across the different splits. 
The non-IPC domains -- Delivery, 
Depots, 
Driverlog, 
Freecell, 
Goldminer, 
Grippers, 
Reward, and Visitall -- start with moderate object/atom counts in training ($\leq70$ objects, $\leq1100$ atoms) and scale to hundreds of objects and up to 6,543 atoms at test time (e.g., Reward and Visitall). 

The IPC-domain 
test splits explode in complexity. For example, Blocksworld grows from at most 20 in the train set to 488 objects in the test set, Miconic's atom counts reach 21,443, Rovers tops 294,553 atoms, and Sokoban stretches to nearly $10^4$ objects.

\end{document}